\documentclass[a4paper,fleqn]{cas-dc}

\usepackage[authoryear,longnamesfirst]{natbib}

\def\tsc#1{\csdef{#1}{\textsc{\lowercase{#1}}\xspace}}
\tsc{WGM}
\tsc{QE}
\def\Dec{\operatorname{Dec}^{\mathcal L}}

\usepackage{booktabs}
\usepackage{float}
\usepackage{morefloats}
\extrafloats{100}
\usepackage{placeins}
\usepackage{wrapfig}
\usepackage{amsmath}
\usepackage{amssymb}
\usepackage{todonotes}
\usepackage{multirow}
\usepackage[table]{xcolor}
\usepackage{graphicx}
\usepackage{svg}
\usepackage{lineno}
\usepackage{makecell}
\usepackage{todonotes}
\usepackage{xcolor}
\usepackage{mathtools}
\definecolor{darkorange}{HTML}{CC8400}
\definecolor{darkred}{HTML}{CC0000}

\usepackage{xurl}
\renewcommand{\paragraph}[1]{\vspace{1em}\noindent\textbf{#1}.}

\newcommand{\vect}[1]{\mathbf{#1}}

\newcommand{\shapecoeff}{\boldsymbol{\beta}}
\newcommand{\shapedim}{{\left| \shapecoeff \right|}}

\newcommand{\posecoeff}{\boldsymbol{\theta}}

\newcommand{\template}{\mathbf{\bar{M}}}
\newcommand{\restpose}{\mathbf{M}}

\newcommand{\blendweights}{\mathcal{W}}

\begin{document}
\let\WriteBookmarks\relax
\renewcommand{\floatpagefraction}{1}
\renewcommand{\textfraction}{0.05}
\renewcommand{\topfraction}{0.9}
\renewcommand{\bottomfraction}{0.9}
\renewcommand{\dbltopfraction}{0.9}
\renewcommand{\dblfloatpagefraction}{1}

\shorttitle{}    

\shortauthors{}  

\title [mode = title]{Mesh-Agnostic Refinement Models for Latent Space Encoding of 3D Human Body Shapes}  


% % Title footnote 1.
% % eg: \tnotetext[1]{Title footnote text}


% \author[1]{}%[<options>]

\author[4,5]{Chaima Zouaoui}
\fnmark[1]
\author[2]{Jonathan Cerqueira}
\fnmark[1]
\author[1]{Sylvain Arguillère}
\author[2]{Martin Bauer}
\author[3]{Nicolas Charon}
\author[4,5]{Mohamed Daoudi}
\author[3]{Emmanuel Hartman}
\fntext[1]{These authors contributed equally to this work.}

% % Corresponding author indication
% \cormark[1]

% % Footnote of the first author

% % Email id of the first author
% \ead{}

% % URL of the first author

% % Credit authorship
% % eg: \credit{Conceptualization of this study, Methodology, Software}


\affiliation[1]{organization={Université de Lille, Département de Mathématiques},
            addressline={Bâtiment M2, Avenue Carl Gauss, Cité Scientifique}, 
            city={Villeneuve-d'Ascq},
%          citysep={}, % Uncomment if no comma needed between city and postcode
            postcode={59655}, 
            state={Nord},
            country={France}}
            
\affiliation[2]{organization={Florida State University, Department of Mathematics},
            addressline={208 Love Building, 1017 Academic Way}, 
            city={Tallahassee},
%          citysep={}, % Uncomment if no comma needed between city and postcode
            postcode={32306-4510}, 
            state={Florida},
            country={United States}}

\affiliation[3]{organization={University of Houston, Department of Mathematics},
            addressline={3551 Cullen Blvd, PGH Room 641}, 
            city={Houston},
%          citysep={}, % Uncomment if no comma needed between city and postcode
            postcode={77204-3008}, 
            state={Texas},
            country={United States}}
            
\affiliation[4]{
    organization={Univ. Lille, CNRS, Centrale Lille,
    Institut Mines-Télécom, UMR 9189 CRIStAL},
    city={F-59000 Lille},
    %postcode={59000},
    country={France}
}       
    \affiliation[5]{
    organization={IMT Nord Europe, Institut Mines-Télécom, Univ. Lille, Centre for Dihal Systems},
    city={F-59000 Lille},
    %postcode={59000},
    country={France}
}

% \author[2]{}%[]

% % Footnote of the second author

% % Email id of the second author
% \ead{}

% % URL of the second author

% % Credit authorship

%\affiliation[2]{organization={},
%          citysep={}, % Uncomment if no comma needed between city and postcode

% \cortext[1]{Corresponding author}

% % Footnote text


\begin{abstract}
Parametric human body models such as the Skinned Multi-Person Linear (SMPL) model provide compact representations of human shape and pose, but their reconstructions remain affected by both errors in estimated latent parameters and limitations in the decoder's expressive capacity. We propose a framework for refining these reconstructions by exploiting the geometric discrepancy between an initial decoded surface and a target mesh, without requiring vertex correspondences between them. The framework comprises three complementary approaches. Latent-VaRefine uses varifold-distance gradients to correct the estimated latent code while preserving the original parametric representation. Vertex-VaRefine processes multiscale varifold-gradient features with a DiffusionNet backbone to predict vertex displacements in a canonical pose, yielding an approximately pose-equivariant refinement pipeline. ElaOpt directly optimizes surface deformations under elastic regularization and normal-direction constraints. On AMASS, Vertex-VaRefine reduces the mean vertex-to-vertex error of OmniFit reconstructions from 1.181 to 0.342~cm, a 71.0\% reduction, while combining Latent-VaRefine with ElaOpt reduces the asymmetric Chamfer distance of ArtEq reconstructions from 1.681 to 0.717~cm, a 57.3\% reduction. Evaluations with SMPL-H and SMPL-X demonstrate the framework's adaptability to different body models, while additional experiments on remeshed targets and raw scans assess its robustness to changes in discretization and geometry beyond the parametric model. These results highlight the complementary strengths of latent-space correction, learned vertex refinement, and elastic optimization, offering different trade-offs between preserving the model parameterization and improving geometric fidelity. Our code is publicly available at
\url{https://github.com/chaimazouaoui9/mesh-agnostic-body-refinement}.

%\textcolor{red}{I don't think this abstract represents our work, especially with the insistence on clothed bodies} We also present qualitative results on clothed human body scans, where clothing geometry and fine facial details are not adequately captured by SMPL, and show a simple example of scan animation using our framework.
\end{abstract}

% \begin{highlights}
% \item 
% \item 
% \item 
% \end{highlights}

%\nocite{*}

% Keywords

\maketitle

% Main text
% \section{}\label{}
\begin{figure*}[]
    \centering
    \includegraphics[width=0.9\linewidth]{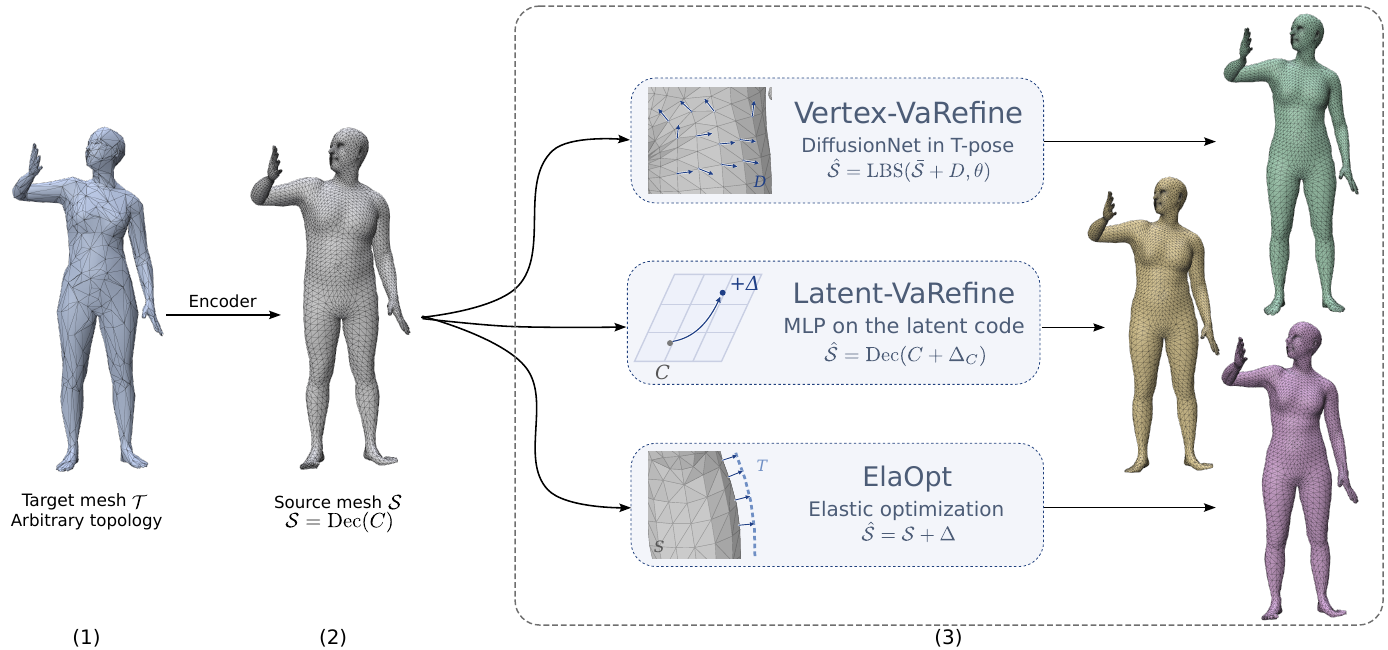}
    \caption{Overview of our refinement framework. (1)~Given a target mesh $\mathcal{T}$ of arbitrary topology, (2)~an existing encoder provides an initial reconstruction, the source mesh $\mathcal{S}=\mathrm{Dec}(C)$, with the fixed topology of the body model. (3)~Our three methods refine this reconstruction: Latent-VaRefine corrects the latent code, Vertex-VaRefine predicts a per-vertex correction in the canonical T-pose that is then re-posed by linear blend skinning, and ElaOpt deforms the mesh vertices by elastic optimization.}
    \label{fig:teaser}
\end{figure*}
\section{Introduction }
\label{sec:intro}
The accurate capture of 3D deformable objects is commonly done through registration of a given target $\mathcal{T}$ (e.g.\ a triangular mesh produced by a 3D scanner) onto a 3D mesh with a fixed, canonical mesh topology. For human body shapes, this is most often done by relying on a parametric representation such as the Skinned Multi-Person Linear model (SMPL)~\citep{loper2015smpl} and its extensions SMPL-H~\citep{romero2017embodied} and SMPL-X~\citep{pavlakos2019expressive}. Such a model is given by a latent space $\mathcal{L}=\mathbb{R}^N$ and a decoder $\mathrm{Dec}^\mathcal{L}:\mathcal{L}\rightarrow\mathbb{R}^{V \times 3}$ which, to a given latent code, associates a triangular mesh with $V$ vertices. Registration of $\mathcal{T}$ then becomes equivalent to \textit{code retrieval}, i.e., to finding the latent code whose decoded mesh best represents the target.

%This also allows for animation of the initial shape 

In recent years, many methods for latent code retrieval have been proposed. % pertaining to the SMPL latent space and decoder. 
 Optimization approaches such as~\citep{bogo2016keepSMPL,pavlakos2019expressive,kanazawa2018ShapePose,feng2021BodiesRegression} require heavy user supervision. Learning-based methods, on the other hand, have recently made notable breakthroughs \citep{IPNetBhatnagar20a}. State-of-the-art frameworks based on equivariant networks such as ArtEq \citep{arteq2023} have been shown to generalize well to unseen poses.  Despite these advances, reconstruction errors can arise from two distinct
sources. First, the retrieval algorithm may fail to identify the parameters
that best explain the observed target, even when its geometry is
representable by the model. Second, the target may contain geometric
details that lie outside the decoder's representational capacity.
These two sources of error call for complementary corrections:
refining the latent code addresses estimation errors, whereas recovering
geometry beyond the model requires additional surface deformations.

However, a limitation of all latent code retrieval methods is their inability to retrieve details not representable by the latent space, such as raw scans, or meshes from other, incompatible databases. Hence, a natural question is how to account for these various extra details without losing advantages intrinsic to the parametric model, such as the possibility of animating a human shape, once registered onto the model, through some evolution in the pose parameters of the parametric model.  A sufficiently accurate initial reconstruction provides more than an
estimate of the target: it also supplies a structured reference on which
the remaining geometric discrepancy can be represented. This motivates
a residual refinement strategy that uses the reconstructed surface as
a target-adapted reference. The central challenge is to encode and exploit
this discrepancy without requiring vertex correspondences between the
source and target, whose sampling and connectivity may differ.
A further challenge for learned corrections is to account for articulation,
so that similar geometric discrepancies can be processed consistently
across different body poses.
%For example, even \textcolor{red}{Talking about clothing here clearly implies that we tackle the issue in the paper, while we (rightfully) put it in the future works section!} simple clothing  cannot be recovered through the SMPL model, nor can new body shapes be reconstructed accurately. For clothed scans, some methods have addressed this issue by applying a deformation field to the generated SMPL reconstruction, resulting in the SMPL+D model \citep{IPNetBhatnagar20a}, but these are highly specialized and only deal with the addition of clothes. 

In this work, we introduce a collection of complementary tools that can be integrated with any latent code retrieval algorithm that produces a decoded surface $\mathcal{S}=\mathrm{Dec}(C)$ approximating a target shape $\mathcal{T}$ within a chosen latent representation. Our approaches exploit the geometric proximity between $\mathcal{S}$ and $\mathcal{T}$ to estimate corrections, either in the latent space or directly on the vertices, leading to a corrected surface $\hat{\mathcal{S}}$ capable of recovering finer geometric details of $\mathcal{T}$. Central to our framework are mesh-agnostic representations of the discrepancies between $\mathcal{S}$ and $\mathcal{T}$, which allow the correction process to train in a way that remains independent of the topology of the target mesh. For the learning-based approaches, we encode the source--target discrepancy
through varifold-distance gradients, evaluated with respect to either
latent parameters or surface vertices. For vertex-wise prediction,
we additionally express these features in a canonical pose and re-pose
the predicted correction, encouraging approximate consistency across
articulations. The optimization-based approach instead fits the target
directly through a geometric objective regularized by elastic deformation
energies. Together, these mechanisms address different requirements:
preserving the original parametric representation, learning transferable
surface corrections, or refining an individual reconstruction without
training.
Our three main contributions are the following:
\begin{itemize}
    \item \textbf{Latent-VaRefine:} a learning-based method that corrects the latent code \(C\in \mathcal{L}\) directly in the latent space, yielding a refined code $C+\Delta_C$.
    \item \textbf{Vertex-VaRefine:} an equivariant, learning-based method that predicts a per-vertex displacement field $\Delta\in\mathbb{R}^{V\times 3}$, in which the mesh-agnostic representation is used directly as a per-vertex signal and processed by DiffusionNet.
    \item \textbf{ElaOpt:} an alternative optimization-based strategy relying on a simple elastic energy formulation that deforms $\mathcal{S}$ toward $\mathcal{T}$ while discouraging unrealistic distortions and retaining reposability.
\end{itemize}
A schematic description of these three methods can be seen in Fig.~\ref{fig:teaser}. They may be used individually or in concert, depending on the objective: if retaining a low-dimensional latent code representation is key, then Latent-VaRefine alone may be best. If, instead, simply acquiring accurate, low-noise meshes is the objective, then Vertex-VaRefine may be more practical. Combinations of the methods may also be useful. For example, as will be seen in our results, the strongest improvement in various error measures came from first applying Latent-VaRefine to obtain a better starting point for optimization and then using ElaOpt. These alternatives expose a practical trade-off between preserving
the original model parameterization and increasing geometric flexibility.
Latent-space refinement retains the decoder's pose and shape
parameterization, whereas vertex-wise refinement introduces additional
degrees of freedom to better approximate the target surface.
The latter retains the source mesh connectivity and can reuse its skinning
structure, but the corrected surface need not correspond to a latent code
of the original model. The appropriate refinement therefore depends on
the representation required by downstream applications, as well as on
reconstruction accuracy.

%Evaluations of all proposed methods, both individually and in combination, are performed against a base, unadjusted state-of-the-art latent code retrieval approach in the context of the SMPL latent space and decoder, as well as its extensions SMPL-H and SMPL-X.
We evaluate the proposed refinement methods, individually and in
combination, on AMASS using ArtEq and OmniFit as initial reconstruction
methods. Vertex-VaRefine reduces the mean vertex-to-vertex error of
OmniFit reconstructions from 1.181 to 0.342~cm, corresponding to a
71.0\% reduction. Starting from ArtEq reconstructions, combining
Latent-VaRefine with ElaOpt reduces the asymmetric Chamfer distance
from 1.681 to 0.717~cm, a 57.3\% reduction. These results illustrate
the complementary strengths of learned vertex refinement for
correspondence accuracy and latent-space refinement followed by
elastic optimization for surface fidelity. Further evaluations
with SMPL-H and SMPL-X demonstrate the adaptability of the framework
to different body models, while experiments on remeshed targets
and raw scans assess its robustness to changes in target
discretization and its ability to refine geometry beyond the
parametric model. 

\section{Background and Related Work}
\label{sec:related}
\subsection{Human Body Latent Spaces}

Statistical parametric body models constitute the dominant paradigm for low-dimensional human body representation. SMPL~\citep{loper2015smpl} introduced a skinned linear representation where shape coefficients $\shapecoeff$ are obtained by PCA over a population of registered scans, while pose coefficients $\boldsymbol{\posecoeff}$ encode joint-angle rotations driving a linear blend skinning deformation. SMPL was later extended to SMPL-H~\citep{romero2017embodied} for articulated hands and further to SMPL-X~\citep{pavlakos2019expressive} for unified body, hand, and facial expression modeling. This was made more sparse in STAR~\citep{osman2020star}. While these parametric models provide a compact and differentiable parameterization, the joint-angle pose space $\boldsymbol{\posecoeff}$ is itself unconstrained: arbitrary parameter values may correspond to implausible meshes. To address this, learned priors over the pose manifold have been proposed, including the VAE-based VPoser~\citep{pavlakos2019expressive}, the GAN-based adversarial pose prior of~\cite{davydov2022adversarial}, and the neural distance field formulation Pose-NDF~\citep{tiwari2022posendf}.
Yet, the expressive power of the latent space in these approaches remains limited by the model's parameterization, and any geometry lying outside this manifold, fine surface detail, and clothing cannot be recovered through latent codes alone. In the present work, we will largely make use of the base SMPL model, which is described in more details below. The methods we propose are, however, not specific to SMPL and apply to any latent-space body model, such as SMPL-H and SMPL-X (see Section~\ref{sec:generalizability}).

% A brief overview and discussion of this model can be found in Section 2 of the supplementary matrerial. 

\paragraph{The SMPL, SMPL-H and SMPL-X Latent Spaces and Decoders}
The Skinned Multi-Person Linear model (SMPL), as introduced in~\citep{loper2015smpl} and its extensions SMPL-H~\citep{romero2017embodied} and SMPL-X~\citep{pavlakos2019expressive} share the same construction. Each consists of latent codes $C=(\shapecoeff,\posecoeff) \in \mathbb{R}^{\shapedim}\times \mathbb{R}^{K \times 3}$ and a decoder that maps any such code to the positions of the $V$ vertices of a mesh, i.e.\ an element of $\mathbb{R}^{V \times 3}$, where the number of vertices $V$ is fixed for each model ($V=6890$ for SMPL and SMPL-H, and $V=10475$ for SMPL-X). The $\shapecoeff$ are shape parameters, the number of which, \(|\shapecoeff|\), can range up to 300 and is commonly set to 10. The pose parameter vector $\posecoeff$ encodes the relative rotation of each of the $K$ articulated joints relative to its parent in the kinematic tree, in axis-angle representation, with $K=23$ for SMPL, $K=51$ for SMPL-H (articulated fingers) and $K=54$ for SMPL-X (additional jaw and eye joints). The model uses PCA to account for variations in body shape, and Linear Blend Skinning (LBS), $W(\cdot)$, to pose the mesh. We denote the resulting maps from latent codes to meshes by $\mathrm{Dec}^{\mathrm{SMPL}}(\cdot)$, $\mathrm{Dec}^{\mathrm{SMPL\text{-}H}}(\cdot)$ and $\mathrm{Dec}^{\mathrm{SMPL\text{-}X}}(\cdot)$, respectively; it is computed in three steps.

A rest-pose mesh is first produced by
\begin{equation}
\label{eq:SMPL_decoder}
     \restpose(C)=\restpose(\shapecoeff, \posecoeff) = \template + B_S(\shapecoeff) + B_P(\posecoeff), 
\end{equation}
where $\template \in \mathbb{R}^{V \times 3}$ is the template mesh, $B_S(\shapecoeff)$ is the linear transformation from shape parameters $\shapecoeff$ to shape displacements, and $B_P(\posecoeff)$ is the linear transformation from pose parameters to blendshape correctives that account for soft-tissue deformations due to pose. 
Next, joint locations are obtained for the rest-pose mesh via the linear joint regressor $J(\shapecoeff) : \mathbb{R}^{\shapedim} \to \mathbb{R}^{(K+1)\times 3}$, $J(\shapecoeff) = (t_0, \dots, t_K)$, where $t_0$ is the root joint. 
Finally, Linear Blend Skinning (LBS), denoted $W(\cdot)$, is applied to the rest-pose mesh with skinning weights $\blendweights$ to obtain the output vertices. For SMPL, for instance, this gives:
\begin{equation}
\label{eq:smpl}
\resizebox{0.9\columnwidth}{!}{$
    \mathrm{Dec}^{\mathrm{SMPL}}(C) = \mathrm{Dec}^{\mathrm{SMPL}}(\shapecoeff, \posecoeff) =W(
    \restpose(\shapecoeff, \posecoeff), J(\shapecoeff), \posecoeff, \blendweights ) %
    $.}
\end{equation}
and $\mathrm{Dec}^{\mathrm{SMPL\text{-}H}}$ and $\mathrm{Dec}^{\mathrm{SMPL\text{-}X}}$ are obtained analogously with their own template, blendshapes, joint regressor and skinning weights.

\subsection{Latent Code Retrieval}
Given a target mesh or point cloud, \emph{retrieving} the latent code that best reconstructs it is a non-trivial inverse problem. Classical optimization-based fitters such as SMPLify~\citep{bogo2016keepSMPL} and SMPLify-X~\citep{pavlakos2019expressive} minimize a 2D/3D keypoint-based energy regularized by a pose prior. Regression-based approaches such as HMR~\citep{kanazawa2018ShapePose} and PIXIE~\citep{feng2021BodiesRegression} predict SMPL latent codes in a single forward pass from images, while IPNet~\citep{IPNetBhatnagar20a} regresses SMPL latent codes from point clouds via implicit part segmentation. Recently, ArtEq~\citep{arteq2023} introduced an articulated $\operatorname{SE}(3)$-equivariant architecture that generalizes the SMPL fitting to unseen poses by enforcing equivariance in the encoder. VariShaPE~\citep{hartman2025varishape} proposed an alternative approach via the VariGrad descriptor, a vertex-wise feature that is robust to changes in mesh connectivity and triangulation from~\citep{hartman2023varigrad}. A common limitation persists across these encoders in that they are constrained by the expressive capacity of the underlying body model. The recovered latent code can only describe geometry that lies within the manifold spanned by the decoder. Our work takes a fixed, chosen encoders' outputs as input and introduces a learned correction module that recovers the residual geometry left out by the latent decoder.

\section{Problem Formulation and Setup}
In this section we will present a rigorous mathematical formulation of the latent code recovery problem.

\subsection{Latent Code Recovery} Consider some \emph{target} mesh \(\mathcal T\) for which one would like to identify the latent code \(\mathcal{C}\) in some latent space \(\mathcal{L}=\mathbb{R}^N\) so that the decoded shape \(\mathcal{S}=\mathrm{Dec}^{\mathcal{L}}(\mathcal{C})\) will best approximate \(\mathcal T\).  Note that, by construction, $\mathcal{S}$ has a fixed mesh topology that is typically determined by the template surface from the decoder, e.g. $\template$ for the SMPL model in Eq.~\eqref{eq:SMPL_decoder}. This generally prevents, outside of artificial examples, the reconstructed $\mathcal{S}$ to match \(\mathcal T\) exactly and often makes the estimation of the optimal latent code a difficult problem to solve in practice. Further, due to the expressive limits of the latent space and decoder, retaining fine details from the target mesh is typically impossible, even with optimal recovery. Our goal is thus to propose different general strategies for refining estimated latent codes or their corresponding meshes.

\subsection{Setup}  In the rest of the paper, the \emph{target} mesh, which represents the shape to be \emph{registered} on by the model will always be denoted $\mathcal{T}$. Our second input is a \emph{source} mesh $\mathcal{S}=\mathrm{Dec}^{\mathcal{L}}(C)$ where \(C\in\mathcal{L}\) is some latent code in the chosen latent-space $\mathcal{L}=\mathbb{R}^N$ and obtained by some (possibly suboptimal) latent code estimation method. The mesh structure of the source is prescribed by the given decoder. In most of our experiments, we focus specifically on the SMPL latent-space and its decoder $\mathrm{Dec}^{\mathrm{SMPL}}$. Our methods are, however, not specific to SMPL, (theoretically, nor are they even restricted to latent spaces representing human body shapes); to illustrate this, we also apply them to two other latent-space models, SMPL-H and SMPL-X (Section~\ref{sec:generalizability}). Our only base assumption is that $\mathcal{S}$ and $\mathcal{T}$ remain sufficiently ``close'' to one-another (for example, in terms of Chamfer distance). In most of our experiments, $C$ is obtained by applying ArtEq \citep{arteq2023}, a state-of-the-art latent code retrieval method, to $\mathcal{T}$.

We propose several strategies which may be combined or used independently to bridge the remaining gap between \(\mathcal{S}\) and \(\mathcal{T}\): (1) a method that corrects the estimated latent code \(C\), and (2) two methods that operate directly on the vertices. Among those, deep learning based methods will require an approach for consistently encoding the relationship between the source \(\mathcal{S}\) and the target shape $\mathcal{T}$. Indeed, unlike \(\mathcal{S}\), $\mathcal T$ is typically not assumed to have a fixed mesh structure. To address this, we shall rely on \emph{VariGrad}, an invariant mesh representation framework based on geometric measure theory, which we introduce in the next section. 

\section{The VariGrad representation}

\label{sec:varigrad}
%\todonicolas{It would make sense to subdivide this section into two or three subsections, rather than just a bunch of paragraphs. I also think that there should be a couple sentences giving some motivation about the equivariance properties being discussed.}
We now recall the definition of the varifold distance and its gradient, VariGrad, which will be an essential ingredient for our learning based approaches. We also explain its equivariant properties and deduce an \emph{approximately pose-invariant} version of VariGrad. As shown in \cite{arteq2023}, this property allows for better generalization of learned methods, which also translate into requiring a smaller training sample. Indeed, theoretically, with perfect pose equivariance, only training on surfaces with a neutral pose would be required.
\subsection{Varifold distance}

\label{varifolddistancedescription}
The family of reproducing kernel varifold distances originates from the representation of shapes, in particular surfaces, as a specific type of geometric measures known as varifolds. We refer to \cite{kaltenmark2017general} for a more complete introduction to these metrics. The key feature of this distance is its (quasi-)invariance to resampling/re-meshing. The following is only a brief description of how they are computed in practice.

For a mesh $\mathbf{V}$, we first compute, for each face $k=1,\ldots,F$, its barycenter $c_k \in \mathbb{R}^3$, its oriented unit normal vector $N_k \in \mathbb{S}^2$ and its area $r_k>0$. Similarly, for a second mesh $\mathbf{V}'$ we denote by $c'_\ell$, $N'_\ell$ and $r'_\ell$ the barycenters, unit normals and areas of its faces $\ell=1,\ldots,F'$. The family of varifold distances we consider in the work are defined as follows:
\begin{align}
    \hspace{-1em}d^{\mathrm{Var}}_\sigma(\mathbf{V},\mathbf{V}')^2
    = {} &\hspace{-0.5em}\sum_{k=1}^{F} \sum_{k'=1}^{F}
      G_\sigma(\|c_k-c_{k'}\|)\, \Gamma(N_k \cdot N_{k'})\, r_k r_{k'} \nonumber\\
    &\hspace{-3em}+ \sum_{\ell=1}^{F'} \sum_{\ell'=1}^{F'}
      G_\sigma(\|c'_\ell-c'_{\ell'}\|)\, \Gamma(N'_\ell \cdot N'_{\ell'})\,
      r'_\ell r'_{\ell'} \nonumber\\
    &\hspace{-4em}- 2\sum_{k=1}^{F} \sum_{\ell=1}^{F'}
      G_\sigma(\|c_k-c'_\ell\|)\, \Gamma(N_k \cdot N'_\ell)\, r_k r'_\ell
      \label{eq:var_dist}
\end{align}
%\textcolor{red}{Also you sometimes just skip lines in text code! This means something (namely a new paragraph is started)!}
where \(G_\sigma:\mathbb{R}_{\ge0}\to\mathbb{R}\) is a spatial kernel, taken throughout to be the Gaussian kernel of width $\sigma$ given by \(G_\sigma(t) = e^{-t^2/\sigma^2}\),  and $\Gamma:[-1,1]\rightarrow \mathbb{R}$ is an orientation kernel. 
%In this paper, we always set.
%and  is a function that defines a kernel over the angle between normal directions. 
%\begin{itemize}
%    \item Unoriented Varifold Distance:
%    Taking $\Gamma(N\cdot N') = (N\cdot N')^2$ in Eq.~\eqref{eq:var_dist} yields the unoriented varifold distance $d^{\mathrm{Var,unor}}_\sigma$. Since this kernel depends only on the square of the inner product between normals, it is invariant to a flip of either mesh's normal orientation.
%    \item Oriented Varifold Distance:
    Taking in Eq.~\eqref{eq:var_dist}, as an orientation kernel, \(\Gamma(N\cdot N')=e^{(N\cdot N'-1)/\tau^2}\) for some angular width $\tau>0$  yields the oriented varifold distance. %$d^{\mathrm{Var,or}}_\sigma$. %Unlike the unoriented variant, 
    This orientation kernel distinguishes aligned from anti-aligned normals. Both the Latent-VaRefine model of Section~\ref{sec:latent_correction} and the multi-scale VariGrad features of Vertex-VaRefine (Section~\ref{ssec:vertex_varefine}) rely on this oriented formulation.
%\end{itemize}

\subsection{VariGrad}
\label{ssec:def_varigrad}
As explained above, we assume that every source mesh $\mathcal{S}$ produced by any latent-space decoder has a fixed, canonical topology. In contrast, we allow the target mesh, as it is typically the output of some scanning acquisition procedure, to have both arbitrary connectivity and an arbitrary number of vertices. %e.g., when it represents a clothed scan or a high-resolution acquisition.
The VariGrad framework of~\cite{hartman2023varigrad} proposes to represent $\mathcal{T}$ via the gradient of the above varifold distance. More specifically, the VariGrad representation of the target mesh $\mathcal{T}$ with respect to the source $\mathcal{S}$ is the vector field (on $\mathcal{S}$) denoted  
\[\mathmakebox[\linewidth][c]{
v\doteq \text{VariGrad}_\sigma(\mathcal{S},\mathcal{T})\in \mathbb{R}^{V\times 3}
}\]
and defined as the gradient of the squared varifold distance between $\mathcal{S}$ and $\mathcal{T}$ with respect to the vertices $x_1,\ldots,x_V$ of $\mathcal{S}$. In other words, 
\[\mathmakebox[\linewidth][c]
{v=(v_j) \in \mathbb{R}^{V\times 3}\text{ with: }v_j = \nabla_{x_j} d^{\text{Var}}_\sigma(\mathcal{S},\mathcal{T})^2.
}\]
which we interpret as a vector field on the reference mesh \(\mathcal{S}\). %(we explained varifold distance above?) Here $d^{\text{Var}}_\sigma(\cdot,\cdot)$ denotes the varifold distance with spatial scale $\sigma>0$ used in several previous works \cite{kaltenmark2017general,charon2020fidelity,hartman2025varishape}. 
As \(d_{\sigma}^{\text{Var}}\) is (quasi-)invariant to resampling and re-meshing of \(\mathcal{T}\),  so is \(\text{VariGrad}_\sigma(\mathcal{S},\mathcal{T})\).

In the present work, we will take advantage of different variants of the original VariGrad model from \cite{hartman2023varigrad} described above. In Section \ref{sec:latent_correction}, following the VariShaPE model developed in \citep{hartman2025varishape}, we shall consider VariGrad representations taken with respect to the latent space coordinate basis of SMPL at \(C\) (where $\mathcal{S}=\mathrm{Dec}^{\mathrm{SMPL}}(C)$), yielding a lower-dimensional representations of $\mathcal{T}$. Second, in Section \ref{ssec:vertex_varefine}, we leverage a multi-scale extension $v = (v^{\sigma_1},\ldots,v^{\sigma_\ell}) \in \mathbb{R}^{V\times 3 \times \ell}$ that concatenates several VariGrads with respect to a common reference $\mathcal{S}$ but with various different spatial scales $\sigma_j$, which better captures both coarse body proportions and finer features.

%An interesting property of $d_{\sigma}^{\text{Var}}$ is its invariance to rigid motions: for any 

\subsection{Equivariance to Rigid Motions}

Let $g=(R,t)\in \operatorname{SE}(3)$ be a rigid motion, with rotation $R$ in $\operatorname{SO}(3)$ and translation $t$ in $\mathbb{R}^3$, and have \(g\) act on any mesh $\mathcal{M}$ with vertices $(x_i)_{i=1}^{V}$ by $g\cdot\mathcal{M} \doteq (Rx_i+t)_{i=1}^{V}$. That is, it acts as a rigid motion of the vertices that leaves the topology (faces and connectivity) unchanged. Then for the \(k\)-th face its barycenter $c_k^\mathcal{M}$, unit normal $N_k^\mathcal{M}$, and area $r_k^\mathcal{M}$ are transformed under $g$ so that 
\begin{equation}\label{changesunderrigid}
\left\lbrace 
\begin{aligned}&c^{g\cdot \mathcal{M}}_k=g(c_k^\mathcal{M})=Rc_k^\mathcal{M}+t, \\
&N^{g\cdot \mathcal{M}}_k=RN_k^\mathcal{M},\\ 
&r_k^{g\cdot \mathcal{M}}=r_k^\mathcal{M}.
\end{aligned}
\right.
\end{equation}
The varifold distance between two meshes $\mathcal{S}$ and $\mathcal{T}$ is known (see \cite{kaltenmark2017general}) to be invariant to the action of a joint rigid motion on the two meshes, that is,

%we note that computing their varifold distance $d_\sigma(\mathcal{S},\mathcal{T})$ using \eqref{eq:var_dist} only depends on the distances between centers of the faces, the inner products of unit normals of the faces, and the areas of the faces. Consequent of the data in \autoref{changesunderrigid}, when any $g=(R,t)\in \operatorname{SE}(3)$ is applied \emph{jointly} to both meshes $\mathcal{S}$ and $\mathcal{T}$ the varifold distance is unaffected. This is since the centers are each displaced by the same $g$ so that their distance is preserved, the normals are each rotated by $R$ which preserves their inner product, and the face areas are unchanged.

% since $\|(Rx+t)-(Ry+t)\| = \|x-y\|$,
% $\langle Ru,Rv\rangle = \langle u,v\rangle$, and face normals transform
% equivariantly under rigid motion of the mesh, $n(g\cdot\mathcal{M}) =
% R\,n(\mathcal{M})$.
%Consequently, the squared varifold distance is $\operatorname{SE(3)}$-invariant:
\begin{equation}
    d^{\text{Var}}_\sigma(g\cdot\mathcal{S}, g\cdot\mathcal{T})
    = d^{\text{Var}}_\sigma(\mathcal{S},\mathcal{T}),
    \qquad \forall\, g \in \operatorname{SE}(3).
    \label{eq:varifold_invariance}
\end{equation}
%\todonicolas{I don't believe we need all the justification to arrive at this equation. The equivariance to rigid motion is a known property of the varifold distance, we can just cite the corresponding work.}
Differentiating identity~\eqref{eq:varifold_invariance} with respect to the vertex coordinates $x_j$ of $\mathcal{S}$, and using the chain rule for the action $x_j \mapsto g(x_j)=Rx_j+t$, shows that the gradient field inherits an equivariant transformation law. Writing
%, for any reference mesh $\mathcal{S}$ and target mesh $\mathcal{T}$, 
$v = \text{VariGrad}_\sigma(\mathcal{T};\mathcal{S})$ and $v' = \text{VariGrad}_\sigma(g\cdot\mathcal{T}; g\cdot\mathcal{S})$, we obtain
\begin{equation}
    v'_j \;=\; R\, v_j, \qquad j = 1,\ldots,V,
    \label{eq:varigrad_equivariance}
\end{equation}
i.e., the VariGrad vector attached to the transformed vertex $Rx_j+t$ is exactly the original VariGrad vector at $x_j$, rotated by $R$ — the translation drops out, as expected for the gradient of a translation-invariant energy. 
%In words, \emph{VariGrad is $\operatorname{SE}(3)$-equivariant}: a joint rigid motion $g=(R,t)$ of the source/target pair rotates every vector of the \emph{VariGrad} representation by the corresponding rotation part $R$, leaving the field otherwise unchanged. \textcolor{orange}{Properly speaking, isn't this \(\operatorname{SO}(3)\) equivariant and translation invariant rather than \(\operatorname{SE}(3)\) equivariant?}

\subsection{Approximate Pose Invariant Representation} \label{sec:invariant_varigrad}

We first heuristically argue that VariGrad is also \emph{approximately equivariant} to a common change of pose. For example, consider two surfaces $\bar{\mathcal{S}}$ and $\bar{\mathcal{T}}$ corresponding to the neutral pose of some human body model. Now, bend the left arm for each surface, obtaining posed surfaces $\mathcal{S}$ and $\mathcal{T}$. For most points on the left arm of either surface, (say, $\bar{x}_j\in \bar{\mathcal{S}}$ and $\bar{y}_j\in \bar{\mathcal{T}}$ respectively) this corresponds to applying a rigid transformation, say $g=(R,t)$: $x_j=R\bar{x}_j+t$, $y_j=R\bar{y}_j+t$. Then, the Gaussian factors that appear in the computation of $\bar{v}=\mathrm{VariGrad}_\sigma(\bar{\mathcal{S}},\bar{\mathcal{T}})$ mean that its value $\bar{v}_j$ at a vertex $\bar{x}_j$ depends mainly on points located in a close neighborhood of this vertex, all of which are transformed by $g$. Combining this fact with the global equivariance of VariGrad to common rigid transformations, this leads us to assert that VariGrad components attached to those points should undergo \textit{approximately} the same rotation: if $v=\mathrm{VariGrad}_\sigma(\mathcal{S},\mathcal{T}),$ then $v_j\simeq R\bar{v}_j$.

In the case where the pose of $\mathcal{S}=\mathrm{Dec}^\mathcal{L}(\beta,\theta)$ is given by joint rotations and Linear Blend Skinning (as is the case for SMPL and its variants), the rigid transformation $g=g_j(\posecoeff)=(R_j(\posecoeff),t_j(\posecoeff))$ is determined by the pose coefficients, so we can write $v_j\simeq R_j(\posecoeff)\bar{v}_j$. This leads us to define the \emph{approximate pose-invariant version} of the VariGrad representation $\phi\in \mathbb{R}^{V\times 3}$ by projecting the raw representation back into the frame of the neutral pose (which we call \emph{canonical coordinates}):
\begin{equation}
    \phi_j\;\doteq\; R_j(\posecoeff)^{\top}\, v_j\left(\simeq R_j(\posecoeff)^{\top}R_j(\posecoeff)\bar{v}_j=\bar{v}_j\right).
    \label{eq:varigrad_frame_projection}
\end{equation}
On those vertices that are moved by a locally rigid transformation during a pose change, $\phi_j\simeq\mathrm{VariGrad}_\sigma(\bar{\mathcal{S}},\bar{\mathcal{T}})$. Applying this reasoning to any \emph{common} pose change in both source and target, we obtain with the projection $\phi$ an approximate pose invariant representation of $\mathcal{T}$ onto $\mathcal{S}$.

To confirm this heuristic reasoning, we numerically validate this approximate pose-invariance in Table~\ref{tab:varigrad_pose_approx}. We select 100 real source and target pairs from AMASS and compute their representation $\mathrm{VariGrad}^0_\sigma$ at the reference pose $\posecoeff=\vect{0}$, obtaining reference values for comparison. We then apply a common pose change to both the source and the target, rotating a random number of joints (up to 10) by a random angle drawn from $[10^\circ,20^\circ]$, $[10^\circ,40^\circ]$ or $[10^\circ,60^\circ]$. Finally, we compute the raw representation $\mathrm{VariGrad}_\sigma$ of each new target pose onto their respective new source pose, and its projection $\phi$ onto canonical coordinates (Eq.~\eqref{eq:varigrad_frame_projection}). We then compute both of their per-vertex relative error with respect to our reference values. We find that the median per-vertex relative error of $\phi$ remains below $10\%$ for all tested intervals for the approximate invariant representation, and is about four times smaller than that of the raw VariGrad. The mean error is noticeably larger than the median. This reflects a minority of vertices located in the blending regions between joints, where linear blend skinning mixes several joint transformations and the pose-corrective blendshapes introduce non-rigid deformations that a single local rotation per vertex cannot capture.
\begin{table}[t]
\centering
\small
\setlength{\tabcolsep}{3pt}
\caption{Numerical validation of the approximate pose-invariance of the canonical-coordinate Varigrad representation $\mathrm{VariGrad}^0_\sigma$ (Eq.~\eqref{eq:varigrad_frame_projection}), on 100 real AMASS source/target pairs. For each pair, a random number of joints (between 1 and 10) are rotated by a random angle within the given interval, identically for both meshes. We report the per-vertex relative error (\%) between the $\mathrm{VariGrad}$ representation recomputed at this new pose and its value at the reference pose $\posecoeff=\vect0$, for both the approximately invariant and raw versions of VariGrad. The reduction is the ratio of the raw median error to the median error between the two versions.}
\resizebox{\columnwidth}{!}{%
\begin{tabular}{c|cc|cc|c}
\toprule
\textbf{Angle interval}
& \multicolumn{2}{c|}{\textbf{Invariant proj.}}
& \multicolumn{2}{c|}{\textbf{Raw VariGrad}}
& \textbf{Reduction} \\
\textbf{(deg)} & mean & median & mean & median & (median) \\
\midrule
$[10,20]$ & 22.99 & \textbf{4.37} & 32.84 & 17.99 & $\times 4.12$ \\
$[10,40]$ & 37.29 & \textbf{7.17} & 53.44 & 27.89 & $\times 3.89$ \\
$[10,60]$ & 49.62 & \textbf{9.98} & 71.66 & 38.50 & $\times 3.86$ \\
\bottomrule
\end{tabular}%
}
\label{tab:varigrad_pose_approx}
\end{table}
%\section{ Latent Space Correction}
\section{Latent-VaRefine: Latent Space Correction via Adaptive VariGrads}
\label{sec:latent_correction}
In this section, we introduce Latent-VaRefine, a learning based latent space correction framework that refines an initial latent code using a VariGrad representation. More specifically, given a target mesh $\mathcal{T}$ and an initial latent code $C \in \mathcal L= \mathbb R^N$, our goal is to estimate a correction vector $\Delta_C \in \mathbb{R}^N$ such that the refined latent space coefficients $\hat{C} = C + \Delta_C$ generate a mesh $\hat{\mathcal{S}}=\mathrm{Dec}^\mathcal{L}(C + \Delta_C)$ that more accurately approximates the target mesh~$\mathcal{T}$.

\begin{figure*}[t]
    \centering
    \includegraphics[width=\linewidth]{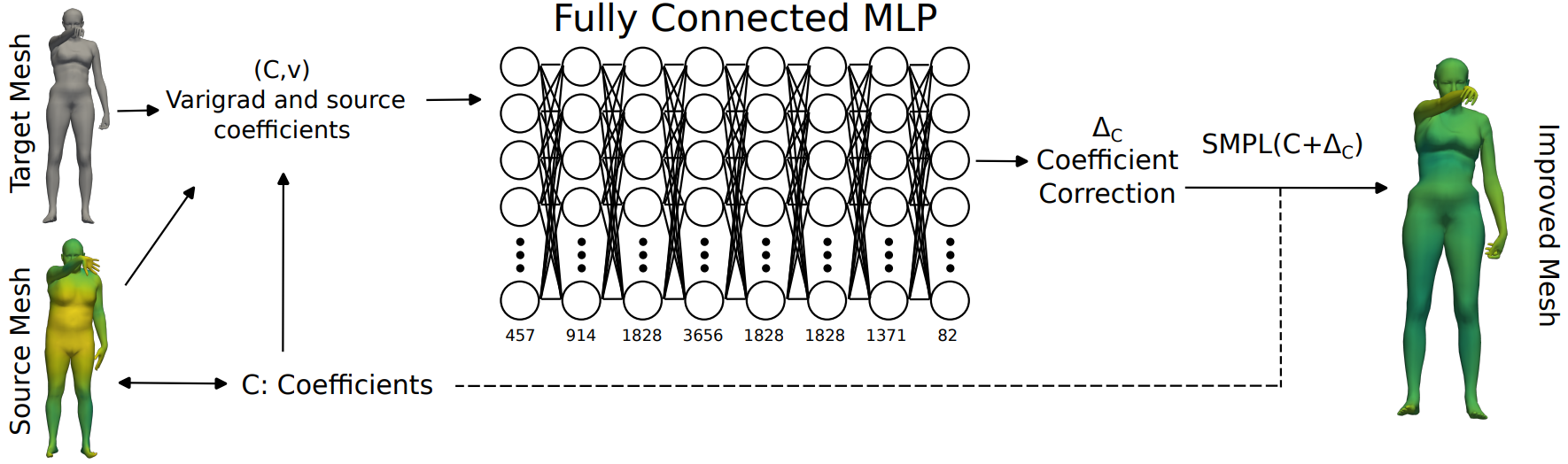}
    \caption{Overview of Latent-VaRefine:  Given a source mesh of fixed SMPL topology $\mathcal{S}$ with coefficients $C$ and a target mesh of arbitrary topology $\mathcal{T}$, we encode the reconstruction error by the low-dimensional VariGrad $v$ with respect to the coefficients $C$ (as defined in the last paragraph of Section \ref{ssec:def_varigrad}). The pair $(C,v)$ is then used to find  $\Delta_C$ via a fully connected MLP. The coefficient correction $\Delta_C$ is then added to the initial coefficients $C$ to produce a latent code $C+\Delta C$ yielding a better mesh reconstruction.}
    \label{fig:varigrad_correction}
\end{figure*}

Our approach is inspired by the VariShaPE framework proposed in~\citep{hartman2025varishape}, which employs the VariGrad representation with respect to the latent-space coordinate basis and a fixed reference mesh~$\mathcal R$ to derive a fixed-dimensional representation of unregistered meshes. This representation can subsequently serve as input to a neural network for learning an embedding into the latent space. A limitation of this representation arises when the target shape differs significantly from the reference mesh, in which case the VariGrad representation may produce suboptimal results. The main idea behind Latent-VaRefine is therefore to apply this approach as a refinement stage on top of an existing latent space encoder. In the examples presented later in Section \ref{sec:implementation_details}, we will use the SMPL representation for human body shapes together with ArtEq \citep{arteq2023} as the primary encoder/code retriever, but the framework is equally compatible with any alternative encoder. 

This two-step process involves the computation of the VariGrad representation relative to the surface $\mathcal{S}=\Dec(C)$ generated by the latent space decoder using the given estimate rather than with respect to a fixed template mesh. Assuming $\mathcal{S}$ is a reasonably accurate estimate of the target $\mathcal{T}$, the anchor point underlying the VariGrad representation is already close to the target mesh. We subsequently use both the latent code estimate $C$ and the corresponding VariGrad representation $v$ as input to a neural network that learns a correction to the estimated latent code, as illustrated in Figure \ref{fig:varigrad_correction}. To ensure that Latent-VaRefine does not produce a worse mesh than we began with, we use a symmetric chamfer measurement to decide whether to keep the resulting refined latent code or return to the original latent code and its corresponding mesh.
While inference with the proposed pipeline can be performed on arbitrary unregistered meshes, training the network requires data with ground-truth latent space coefficients. In our experiments we use data from the AMASS dataset, for which we have access to those ground truth SMPL coefficients.

This type of training data enables us to train the network using a loss function that combines the mean vertex error between the corrected mesh and the ground-truth mesh with a latent-space error term, i.e., the discrepancy between the estimated latent-space parameters and the corresponding ground-truth parameters. 

\section{Vertex-wise Correction}
In this section, we present two different Vertex Correction Frameworks that refine an initial source mesh $\mathcal{S}=\Dec(C)$ reconstructed from a latent code estimate $C$ by using a geometry-aware deformation field to better match the target mesh $\mathcal{T}$. Unlike the previous latent space correction scheme, the main objective is here to allow the reconstruction of retrieve geometric details outside of the latent model. First, we introduce Vertex-VaRefine, a learning-based approach based on an equivariant neural network model. Second, we consider an alternative optimization-based approach involving elastic energies and second-order surface metrics as regularizers. 
In the following, we denote by $(x_1,\dots,x_V)\in \mathbb{R}^{V\times 3}$ the vertices of the source mesh $\mathcal{S}$. The objective is to predict a per-vertex displacement field that we shall write $\boldsymbol{\Delta} \in \mathbb{R}^{V \times 3}$ so that the refined mesh $\hat{\mathcal{S}}$, with vertices $\hat{x}_i=x_i+\Delta_i$, results in a reconstruction closer to $\mathcal{T}$.

\subsection{Vertex-VaRefine: An Equivariant Learning Based Approach for Vertex Correction}

\label{ssec:vertex_varefine}

\begin{figure*}[]
    \centering
    
   \includegraphics[width=1\linewidth]{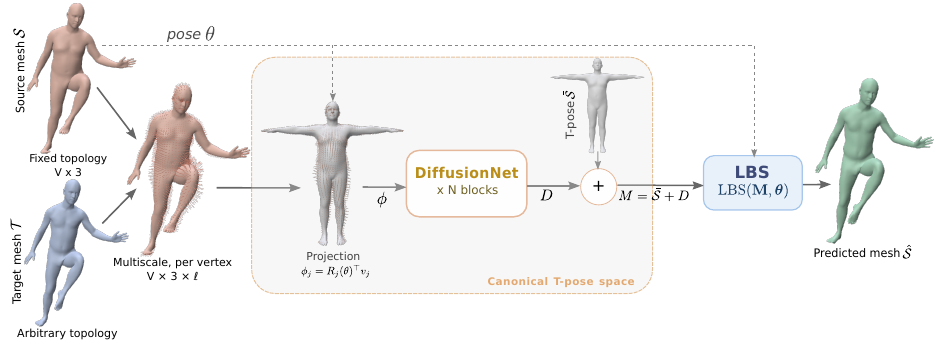}
   \caption{Overview of Vertex-VaRefine. Given a source mesh $\mathcal{S}=\Dec(C)$ of prescribed topology and a target mesh $\mathcal{T}$ of arbitrary topology, we compute multi-scale VariGrad descriptors $v\in\mathbb{R}^{V\times3\times\ell}$ on $\mathcal{S}$ and project them onto the per-vertex rest-pose frames, $\phi_j=R_j(\posecoeff)^\top v_j$, giving approximately pose-invariant features on the neutral-pose shape $\bar{\mathcal{S}}$. A DiffusionNet backbone ($N$ blocks of learned diffusion, spatial gradient features and a shared pointwise MLP) maps $\phi$ to a displacement $D\in\mathbb{R}^{V\times3}$ in T-pose space. The corrected shape $\restpose=\bar{\mathcal{S}}+D$ is re-posed with the source pose, $\hat{\mathcal{S}}=\mathrm{LBS}(\restpose,\posecoeff)$.}
    \label{fig:proposed_model}
\end{figure*}

Figure~\ref{fig:proposed_model} illustrates the overall architecture of our proposed
deformation refinement framework. It combines three key components:

\paragraph{Multiscale VariGrad:} We choose a set of scales $\sigma_1,\dots,\sigma_\ell$ and compute, for each $\sigma_j$, the VariGrad descriptors 
\[\mathmakebox[\linewidth][c]{
v_j=\mathrm{VariGrad}_{\sigma_j}(\mathcal{T};\mathcal{S})=(v_{j,1},\dots,v_{j,V})\in \mathbb{R}^{V\times 3}
}\]
from the source to the target (see Section \ref{ssec:def_varigrad}), which encode the geometric differences between the two shapes. We obtain a total of $3\times \ell$ features on each vertex by concatenation: $v=(v_1,\dots,v_\ell)\in\mathbb{R}^{V\times 3 \times \ell}$.

Once the source--target residual $v$ is encoded vertex-wise, we need a network that can propagate information across the surface while respecting its intrinsic geometry. Standard MLPs treat vertices independently and thus ignore neighborhood structure, while spectral graph convolutions are computationally expensive and basis-dependent. 
% We chose to combine SpiralNet++ layers to which we append a Neural Jacobian fields Module, but other options such as DiffusionNet ~\cite{SharpDiffusionNet} or PoissonNet ~\cite{maesumi2025poissonnet} could also be valid alternatives.

We chose DiffusionNet~\citep{sharp2022diffusionnet} as the backbone that propagates the encoded discrepancy across the surface, but other options such as a SpiralNet++~\citep{gong2019spiralnet++} encoder, or PoissonNet~\citep{maesumi2025poissonnet}, could also be valid alternatives (see Section~\ref{sec:ablation_backbones} and Appendix~\ref{app:backbones}).

\paragraph{DiffusionNet}
DiffusionNet is a general-purpose architecture for deep learning on 3D surfaces, introduced by ~\cite{sharp2022diffusionnet}. Its central idea is that a simple learned diffusion layer provides an effective mechanism for spatial communication across a surface: at each block, per-vertex features are diffused for a learned time scale, so that the spatial support is optimized as a continuous network parameter ranging from purely local to fully global, removing the need to hand-pick neighborhood sizes. Each DiffusionNet block combines this diffusion step with spatial gradient features that support directional (anisotropic) filters and a spatially shared pointwise MLP applied independently at every vertex, and several identical blocks are stacked to form the network. Because diffusion is expressed through intrinsic surface operators (the Laplace–Beltrami operator and mass matrix), which can be discretized on triangle meshes as well as on point clouds~\citep{Sharp:2020:LNT}, the resulting networks are robust to changes in resolution and sampling of a surface. Rather than overfitting to a particular mesh connectivity, the model learns a representation-agnostic solution and can even be trained on one representation, such as a triangle mesh, and evaluated on another, such as a point cloud. This is achieved without any explicit surface convolutions or pooling hierarchies, which makes the architecture simple, efficient, and well suited to our setting.

\paragraph{Pose-Equivariant Prediction and Re-posing}
%All three backbones considered in this section (DiffusionNet, PoissonNet, and SpiralNet++, see the ablation study in the section \ref{sec:ablation_backbones}) share the same mechanism for achieving pose equivariance, and differ only in how the per-vertex correction is decoded from the processed features. 
%\textcolor{red}{Using the same letter as the target, even in a different font, can be confusing.}
Rather than predicting the correction directly on the posed source mesh, we predict it on the canonical neutral pose $(\posecoeff=0)$ shape $\bar{\mathcal{S}}$ obtained from the SMPL shape blendshapes of $\mathcal{S}$, and recover the final displacement by re-posing through linear blend skinning. Concretely, the VariGrad field $v$ computed on the posed source $\mathcal{S}$ is first projected onto the per-vertex rest-pose frame $R_j(\posecoeff)$ obtained from linear blend skinning (Eq.~\eqref{eq:varigrad_frame_projection}), yielding a feature field $\phi=(\phi_1,\dots,\phi_V)$ that is, as discussed in Section~\ref{sec:invariant_varigrad}, approximately invariant to the pose $\posecoeff$. The backbone (here, DiffusionNet) processes $\phi$ into a corrected neutral pose shape $\restpose = \bar{\mathcal{S}} + D$, and the final prediction is obtained by re-posing $\restpose$ with the same pose parameters $\posecoeff$ from the source shape $\mathcal{S}$, in other words:
\begin{equation}
    \hat{\mathcal{S}} = \mathrm{LBS}(\restpose ,\posecoeff)
    = \mathrm{LBS}(\bar{\mathcal{S}}+D,\posecoeff).
    \label{eq:tpose_repose}
\end{equation}

Since both $\phi$ and $D$ live in the canonical T-pose frame, and pose is only reintroduced via the fixed $\mathrm{LBS}$ skinning map, the correction pipeline is by construction \emph{approximately} pose-equivariant: applying it to the same subject in a different pose $\posecoeff'$ yields approximately the same T-pose correction $D$, simply re-posed through Eq.~\eqref{eq:tpose_repose}. This approximate equivariance arises because $\phi$ is itself only approximately pose-invariant (Eq.~\eqref{eq:varigrad_frame_projection}), a discrepancy that is quantified numerically in Table~\ref{tab:varigrad_pose_approx} and whose practical impact on the trained model we isolate via an ablation in Section~\ref{sec:ablation_equivariance}.

\subsection{ElaOpt: An optimization Based Approach Using Elastic Energies}
%\textcolor{Red}{Completely abrupt without any intro what we are trying to do here!}
\begin{figure}
\centering

\includegraphics[width=\linewidth]{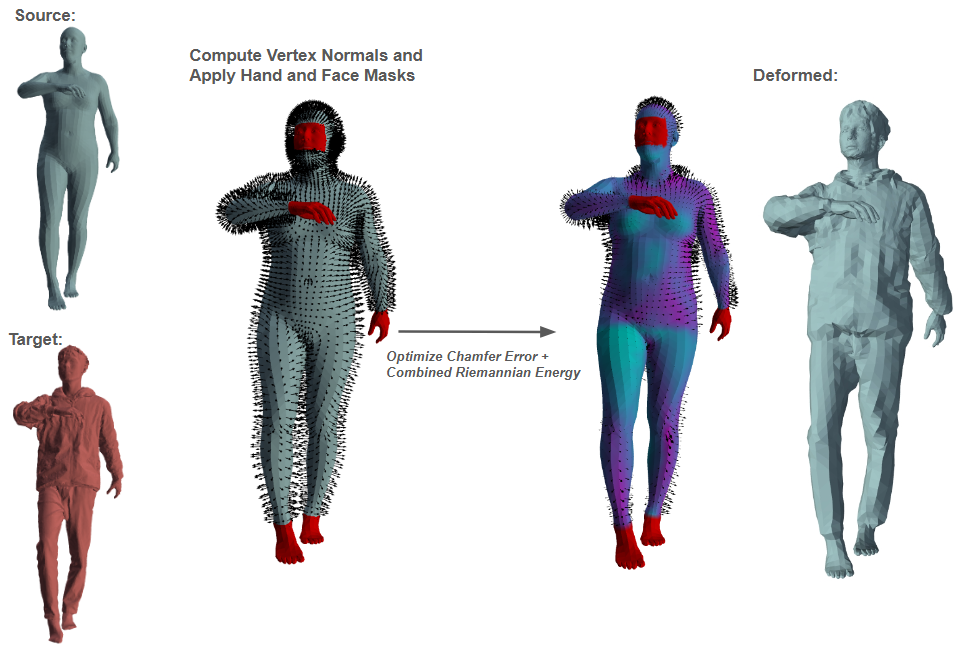}
\caption{Schematic of the optimization based approach for vertex based mesh correction. The correction vector field is estimated by minimizing the sum of a similarity and a regularization energy.}
\label{fig:opt_framework}
\end{figure}

Finally, we introduce ElaOpt, an optimization-based vertex-wise correction model with regularization based on an elastic deformation energy. The novelty of ElaOpt lies in the combination of three components: (i) a mesh-agnostic elastic registration term rooted in mathematical shape analysis; (ii) restricting the admissible deformations to the surface-normal direction; and (iii) an adaptive mask that enforces stronger regularization in high-curvature or noise-prone regions, such as the face and hands. Given a source shape $\mathcal S$ and a target shape $\mathcal T$, we seek a correction field $\boldsymbol{\Delta}$ that minimizes the energy
\[\mathmakebox[\linewidth][c]{
\mathcal E(\boldsymbol{\Delta})
=
\mathcal E^{\operatorname{sim}}(\mathcal S+\boldsymbol{\Delta},\mathcal T)
+
\mathcal E^{\operatorname{reg}}(\mathcal S,\boldsymbol{\Delta}).
}\]
The first term, $\mathcal E^{\operatorname{sim}}$, measures the similarity between the corrected source surface $\mathcal{S}+\boldsymbol{\Delta}$ and the target surface $\mathcal T$, while the second term, $\mathcal E^{\operatorname{reg}}$, regularizes the deformation field $\boldsymbol{\Delta}$. Note that the regularization term depends not only on the correction vector $\boldsymbol{\Delta}$, but takes into account the geometry of the source $\mathcal S$.  We further emphasize that the correction field is constrained to share the mesh structure of the source surface, whereas the target surface may possess a different triangulation. Consequently, the similarity measure must compare surfaces with non-matching mesh discretizations, as is the case for e.g. the Chamfer distance. 

In the following, we provide a high-level description of the chosen energy terms, while full details of the precise energy formulation are deferred to \ref{ELaOptInternals}. We employ the Chamfer distance to define the similarity energy due to its simplicity and computational efficiency, although alternative geometric shape metrics, such as the varifold distance from Section \ref{sec:varigrad}, could also be used. For the regularization term, we draw inspiration from elastic shape analysis. Specifically, we combine a reparameterization-invariant Sobolev $H^2$ Riemannian metric~\citep{hartman2023elastic}---defined using the geometry of the source surface $\mathcal S$---with the Square Root Normal Field (SRNF) distance~\citep{jermyn2012elastic,kurtek2012riemannian,bauer2021numerical} between the source surface and its vertex-corrected counterpart. An ablation study investigating the influence of the various components of the energy is provided in \ref{termwiseElaOptTest}. 
%\label{smplDandElaOpt}
%\label{termwiseElaOptTest}
% In addition, to the regularization energy described above, we employ a hybrid deformation model, where we identify certain regions on which allowable deformations are more constrained. Specifically, vertices of $\mathcal{S}$ are divided into masked (namely hands/feet/face) and unmasked regions. To prevent to drastic deformations in the masked regions we employ either an additional penalty on the $L^2$-norm of the correction vector field in these regions, or we even constrain it to be purely a local rigid deformation on each masked region. 

In addition to the regularization energy described above, we employ a hybrid deformation model in which specific regions are subject to additional deformation constraints. To this end, the vertices are partitioned using a mask $m\in \{0,1\}^V$ into masked regions (e.g. hands, feet, and face for human body shapes) on which $m_i=0$ and unmasked regions  where $m_i=1$. To prevent excessive deformations within the masked regions, we impose an additional penalty on the $L^2$-norm of the correction vector field restricted to these regions.

Furthermore, within the unmasked regions, we restrict the correction vector field to act purely in the normal direction of the source surface, i.e.,
$
\boldsymbol{\Delta}_i = m_i\alpha\, N^{\mathcal S}_i+(1-m_i)\delta_i,\  i=1,\dots,V,
$
where $N^{\mathcal S}$ denotes the normal vector field of $\mathcal S$ and $\alpha$ is a scalar field defined on the surface, while $\delta$ fits the chosen constraints on the masked regions. This additional restriction was seen to promote smoother and more regular corrected surfaces. The unconstrained optimization problem itself is tackled with the L-BFGS method. An schematic of the full pipeline is shown in Figure~\ref{fig:opt_framework}.

%\section{BASELINES}
\section{Implementation Details}
\label{sec:implementation_details}
%\subsection{Evaluation Protocol}
%\label{sec:results_protocol}

\subsection{Baseline Encoder}
As a baseline encoder we employ the state-of-the-art ArtEq model. Since the original ArtEq framework reconstructs meshes with the SMPL male model and is trained on meshes constructed with the male SMPL model, we re-trained ArtEq with the SMPL neutral model to support both male and female body shapes and ensure compatibility with our datasets. To verify that our reported gains do not simply stem from the additional training data used by the two VaRefine models, we also report a second baseline, denoted ArtEq$^{+}$, which is trained on the same data as the full ArtEq + VaRefine pipeline with the SMPL neutral model. 
%\textcolor{red}{I am not sure I understand the difference between Arteq and Arteq+, aren't we training with the neutral models in both cases?}\textcolor{blue}{ Yes, both with neutral model. I added that now.} 
We emphasize that our methods are not dependent on these specific choices of latent space and encoder, but could be employed on top of any others as well. To illustrate this, we use a converter that maps SMPL codes to corresponding SMPL-H or SMPL-X representations (Appendix~\ref{app:converter}). It allows us to transpose both the ground-truth codes of the targets and the ArtEq predictions into these other latent spaces, and thus to run the same experiments on those. We also use the very recently proposed OmniFit~\citep{cai2026omnifit} as an alternative choice of baseline encoder to find initial SMPL-X latent codes (Section~\ref{sec:omnifit}).

\subsection{Latent-VaRefine} 
\label{imp_latent_varefine}
The first layer of the model consists of the computation of VariGrad between the given source-target pair. Here, we specifically choose a single fixed spatial scale of $\sigma = 0.75$cm with an oriented Gaussian kernel as mentioned in Section \ref{varifolddistancedescription} above. Furthermore, the VariGrad is here computed with respect to the full \(375\)-SMPL latent parameters rather than by-vertex: this yields a smaller vector in $\mathbb{R}^{375}$ instead of \(\mathbb{R}^{3\times6890}\) for the vertex-wise version. The obtained VariGrad is then concatenated with the latent code \(C\) from the baseline encoder output. For ArtEq this consists of 82 total parameters: 3 global orientation parameters, 69 pose (\(\posecoeff\)) parameters, and 10 shape (\(\shapecoeff\)) parameters.  In total, this concatenated input is a \(457\) dimensional vector. A fully connected MLP with output layers of dimensions \(\{914, 1828, 3656, 1828, 1828, 1371, 82\}\) with RELU activation functions then performs the inference, where the \(82\)-dimensional output vector is the parameter correction \(\Delta_C\).

Training is performed using a weighted sum of two loss terms. The first is a mean squared error distance between the refined mesh \(\hat{\mathcal{S}}=\mathrm{Dec}^{\mathrm{SMPL}}(C+\Delta_C)\) and the target mesh \(\mathcal{T}\):
\[
\mathcal{L}_\mathrm{MSE}(\hat{\mathcal{S}},\mathcal{T})
=
\frac{1}{V} \sum_{i=1}^{V} \left\| \hat{\mathbf{x}}_i - \mathbf{y}_i \right\|_2^2.
\]
The second is a weighted sum of squared latent parameter errors. Here $\bar{C}$ are the ground truth coefficients.
\begin{align*}
    \mathcal{L}_{\mathrm{SMPL}}(\bar{C},C+\Delta_{C})&=w_{\text{ori}}\sum_{i=1}^3(\bar{C}_{i}-(C_{i}+\Delta_{Ci}))^2\\
    &+ w_{\text{pos}}\sum_{i=4}^{72}(\bar{C}_{i}-(C_{i}+\Delta_{Ci}))^2\\
    &+ w_{\beta}\sum_{i=73}^{82}(\bar{C}_{i}-(C_{i}+\Delta_{Ci}))^2
\end{align*}

We take \(w_{\text{ori}}=w_\beta=1\) and \(w_{\text{pos}}=3\). What remains to set is the relative weighting of MSE and latent code parameter loss. We take a weight of \(w_{\text{MSE}}=10000\). Loss is then given by 
\[
\mathcal{L}(\Delta_C)=w_{\text{MSE}}\cdot\mathcal{L}_\mathrm{MSE}(\mathcal{\hat{S},\mathcal{T}})+\mathcal{L}_{\mathrm{SMPL}}(\bar{C},C+\Delta_C).
\] 
Training is performed for 180 epochs using the Adam optimizer with a learning rate of \(10^{-3.5}\), a batch size of 3000, and a learning rate decay factor of .1 every 60 epochs.

\subsection{Vertex-VaRefine}
\label{imp_vertex_varefine}
At the first stage, multi-scale VariGrad descriptors are computed between the posed source and target meshes using 3 scales: $3$cm, $6$cm and $10$cm. Each vertex-wise VariGrad vector is then projected onto the per-vertex rest-pose frame obtained from linear blend skinning of the pose $\posecoeff$ (Eq.~\eqref{eq:varigrad_frame_projection}), producing a $9$-dimensional, approximately pose-invariant feature $\phi_j$ per vertex of the canonical T-pose shape $\bar{\mathcal{S}}$. This feature field is processed by a DiffusionNet backbone composed of $4$ diffusion blocks of width $128$: each block combines a learned-time heat diffusion step -- parametrized by $k=128$ Laplace--Beltrami eigenfunctions of the T-pose SMPL template, precomputed once and cached across the dataset -- with spatial gradient features and a pointwise MLP shared across vertices. A final linear layer maps the $128$-dimensional per-vertex features to a $3$-dimensional per-vertex displacement $D$, predicted \emph{in T-pose space}. The corrected T-pose shape $\restpose=\bar{\mathcal{S}}+D$ is finally re-posed by linear blend skinning with the same pose $\posecoeff$ (Eq.~\eqref{eq:tpose_repose}) to produce the refined reconstruction $\hat{\mathcal{S}}$. Training is performed for $300$ epochs using the Adam optimizer with a learning rate of $4\times10^{-4}$, batch size $4$, gradient-norm clipping at $1.0$, and a learning rate decay factor of $0.9$ every $50$ epochs. The loss function combines a vertex reconstruction term and an intrinsic gradient regularization term weighted by $\lambda_v=1$ and $\lambda_g=0.08$.

Vertex-VaRefine is trained on a database with point-to-point correspondences between the targets (or ground truths) and the source meshes. As a result, we use the following loss: 
\[\mathmakebox[\linewidth][c]{
\mathcal{L}= \frac{\lambda_v}{3} \,  \mathcal{L}_{\operatorname{MSE}} + \lambda_g \, \mathcal{L}_{\operatorname{grad}} ,
}\]
where $\lambda_v = 1$ and $\lambda_g = 0.08$ are weighting coefficients balancing the vertex reconstruction term and the intrinsic gradient regularization term, respectively. Here $\mathcal{L}_{\operatorname{MSE}}$ is the same Mean Squared Error as in~\ref{imp_latent_varefine}, and the factor of $3$ accounts for the $3$ coordinate channels so that both terms are normalized as a mean over individual scalar entries. $\mathcal{L}_{\operatorname{grad}}$ is the area-weighted average of the squared Euclidean distance between the discrete gradient of the predicted mesh and those of the ground truth i.e.:
\[\mathmakebox[\linewidth][c]{
\mathcal{L}_{\operatorname{grad}} =
\frac{1}{3 \sum_{k=1}^{F} r_k} \sum_{k=1}^{F} r_k \left\|  (\nabla \hat{x})_k - (\nabla y)_k \right\|_{2}^2,
}\]
where $\hat{x}$ and $y$ denote the vertices of the predicted mesh $\hat{\mathcal{S}}$ and of the ground-truth mesh, respectively, and $F$ denotes the number of mesh faces. The operator $\nabla$ denotes the intrinsic discrete mesh gradient: it is computed face-wise, with two vector components per face expressed in a local tangent basis, so that $(\nabla \hat{x})_k$ and $(\nabla y)_k$ are the $k$-th of these $F$ components and $r_k$ is the area of the corresponding face. This term regularizes the deformation at the differential level by matching intrinsic surface gradients. Together, these two losses encourage both accurate and geometrically smooth mesh deformations.

\subsection{ElaOpt}
\label{ELaOptInternals}
The correction field \(\Delta\) is constrained to act in the normal direction of the source surface, i.e.
\[\boldsymbol{\Delta}_i=\alpha_i N^{\mathcal S}_i,
\qquad i=1,\dots,V,\]
where \(N^{\mathcal S}_i\) denotes the unit normal vector of the source surface at vertex \(i\), and \(\alpha\) is a scalar field defined on the surface. The optimization energy is given by
\[\mathcal E(\boldsymbol{\Delta})=
\mathcal E^{\operatorname{sim}}(\mathcal S+\boldsymbol{\Delta},\mathcal T)
+
\mathcal E^{\operatorname{reg}}(\mathcal S,\boldsymbol{\Delta}),\]
where the similarity term is defined using the Chamfer distance
\[\mathcal E^{\operatorname{sim}}(\mathcal S+\boldsymbol{\Delta},\mathcal T)
=
\lambda\,d_{\mathrm{Chamfer}}(\mathcal S+\boldsymbol{\Delta},\mathcal T),\]
and the regularization energy is given by
\[\begin{aligned}
\mathcal E^{\operatorname{reg}}(\mathcal S,\boldsymbol{\Delta})
={}&
\mu_{\mathrm{hf}}
\sum_{i\in \mathcal M_{\mathrm{hf}}}
\|\boldsymbol{\Delta}_i\|^2
+
\mu_{\mathrm{fa}}
\sum_{i\in \mathcal M_{\mathrm{fa}}}
\|\boldsymbol{\Delta}_i\|^2
\\[0.3em]
&+
\frac{1}{10^8}
\Bigl(
\|\boldsymbol{\Delta}\|_{H^1(\mathcal S)}^2
+
\|\boldsymbol{\Delta}\|_{H^2(\mathcal S)}^2
\Bigr)
\\[0.3em]
&+
\frac{1}{10^6}
\left\|
Q(\mathcal S+\boldsymbol{\Delta})
-
Q(\mathcal S)
\right\|_{L^2}^2 ,
\end{aligned}\]
where \(Q(\cdot)\) denotes the Square Root Normal Field (SRNF) representation~\citep{jermyn2012elastic,kurtek2012riemannian,bauer2021numerical}, $\|\cdot\|_{H^k(\mathcal S)}^2$ for $k\in\{1,2\}$ denotes the Sobolev norms calculated w.r.t. the induced surface metric of $\mathcal S$~\citep{hartman2023elastic}, and \(\mathcal M_{\mathrm{hf}}\) and \(\mathcal M_{\mathrm{fa}}\) denote masked regions (of e.g. the hands/feet and face) with corresponding weights $\mu_{\mathrm{hf}}$ and $\mu_{\mathrm{fa}}$ respectively.

To improve optimization stability, the energy is minimized iteratively using a L-BFGS method and a  continuation strategy with increasing weights $\lambda \in \{10^{-8},\,10^{-6},\,10^{-4}\}$,
where the solution obtained for one value of \(\lambda\) is used to initialize the subsequent optimization stage.
%
%\textcolor{red}{\textbf{Belongs elsewhere?:}}
%We provide additional qualitative results in the \textbf{supplementary material} in the form of 4D animated sequences of clothed meshes generated by the ElaOpt method. These animations demonstrate the temporal consistency of the reconstructed geometry as well as the realistic deformation behavior of garments under complex body motions. Four representative examples are included as animated GIFs, showcasing a variety of poses, motion dynamics, and clothing types.

\section{Results}
\label{sec:results}
\subsection{Datasets} 
We use meshes reconstructed by a latent-space retrieval method as source shapes, while the corresponding ground-truth meshes are used as targets. More specifically, given a target mesh $\mathcal{T}$, we first obtain an initial reconstruction using one of the state-of-the-art latent-space body retrieval algorithm ArtEq. This gives a latent code $C=(\shapecoeff,\posecoeff)$, which is in turn used to generate a source mesh $\mathcal{S}=\mathrm{Dec}^{\mathrm{SMPL}}(C)$. The objective of the proposed methods is to predict a code correction $\Delta_C$ and/or a deformation field $\Delta$ along $\mathcal{S}$ that refines it toward $\mathcal{T}$.

To this end, we first train ArtEq on the DFAUST dataset~\citep{AMASS_DFaust}, following the protocol from the original paper \citep{arteq2023}. Among our proposed methods, both VaRefine algorithms require training. For this, we use the BMLmovi dataset~\citep{AMASS_BMLmovi}, a subset of AMASS~\citep{AMASS:ICCV:2019} similar to DFAUST. For each mesh $\mathcal{T}_i$ in BMLmovi, we first extract a latent code $C_i$ through the trained ArtEq network, which corresponds to a source mesh $\mathcal{S}_i.$ Since AMASS contains a very large number of frames, we perform frame sampling to reduce training and testing set sizes. More precisely, we randomly select five frames from each motion sequence. As a result, the final training set contains 9005 target meshes $\mathcal{T}_i$, 9005 SMPL codes $C_i$ and their corresponding decoded source meshes $\mathcal{S}_i$, $i=1,\dots,9005$.

For evaluation, we perform a large-scale evaluation on (randomly selected frames of) all of AMASS, excluding the BMLmovi dataset which was used for training. We also give separate results for multiple AMASS sub-datasets in order to assess the generalization capability of our methods across diverse motion distributions and capture conditions. The final evaluation set contains 80\,072 meshes. For each sample target $\mathcal{T}$ in the dataset, we first apply the trained ArtEq network to obtain a latent SMPL code $C$ and corresponding mesh $\mathcal{S}$, then apply/combine each of the three methods in order to compute a correction to the code $C$ and/or each vertex position of $\mathcal{S}$, then compare the V2V and asymmetric Chamfer distance between the corrected prediction and $\mathcal{T}$.

%\textcolor{red}{We do not use the cape dataset anymore except in limitations/future works!}
%To further evaluate the robustness of the proposed framework for clothed human reconstruction, we conduct additional experiments using the CAPE dataset~\citep{CAPE:CVPR:20} with both the Vertex-VaRefine model and ElaOpt, either of which allow leaving of the SMPL latent space. CAPE provides dynamic 3D clothed human meshes together with minimally clothed SMPL body meshes in point-to-point vertex correspondence, making it well suited for deformation-based mesh refinement. The dataset contains realistic garment deformations from clothing such as shirts, jackets, skirts, and shorts, resulting in substantially more challenging reconstruction scenarios than minimally clothed body datasets. We adopt a subject-independent% and clothing-independent evaluation protocol, reserving unseen subjects and the \textit{shortshort} clothing category exclusively for testing. In total, 8,674 meshes are used for training. Additionally, we qualitatively evaluate the ElaOpt method on clothed meshes from the the THumans dataset~\citep{tao2021function4d} which has high fidelity scans of clothed humans as well as SMPL approximations of the pose and general shape of the scan. 

\subsection{Computational Cost}
%\todomartin{We should report something on both training and inference cost!}
In Table~\ref{tab:computational_cost} we present the computational cost of the proposed refinement frameworks in terms of model complexity and inference efficiency. All experiments were conducted on a machine equipped with an NVIDIA RTX 5000 Ada Generation laptop GPU. For training measurements, the gradient descriptors are assumed to be precomputed. In addition to the reported training cost, the learning based methods required the pre-computation of VariGrad representations for the entire training set. For Latent-VaRefine, where VariGrad representations are computed in the SMPL parameter space using a single scale, this step required 108.48 minutes to compute the descriptors for the full dataset of 89\,077 meshes. In contrast, Vertex-VaRefine computes VariGrad descriptors on mesh vertices using 3 scales, resulting in a preprocessing time of 118.77 minutes for the same dataset. Vertex-VaRefine additionally pre-computes, for each training sample, the T-pose shape, the frame-projected VariGrad features, and the intrinsic gradient and mass operators used by the gradient regularization term $\mathcal{L}_{\operatorname{grad}}$; this step took 2.43 minutes for the
9\,005 training meshes.
%\todosylvain{Ok, I do think varigrad should be included for inference? since that is something that needs to be done if given a new mesh/dataset.}
%\todomartin{I agree, I would only report the inference +Gradient time. The other one is rather meaningless in my opinion.}
\begin{table}[H]
\centering
\small
\setlength{\tabcolsep}{4pt}
\renewcommand{\arraystretch}{1}
\caption{Computational cost of the proposed refinement frameworks. We report the number of trainable parameters, the average training time per epoch, and the average inference time per sample including VariGrad calculations.}
\begin{tabular}{lccc}
\toprule
\textbf{Method} 
& \makecell{\textbf{\#Params} \\\textbf{(M)} }
& \makecell{\textbf{Train}\\\textbf{Time/Epoch(s)}}
& \makecell{\textbf{Inference} \\\textbf{(ms)}}\\% & \textbf{Infer+Grad (ms)}\\
\midrule

Latent-VaRefine 
& 21.43
& 8.2 
& 73.1\\%& 0.097 &

Vertex-VaRefine 
& 0.46
& 260 & 82.4 \\
%& 90 & \text{\textasciitilde}170\\

ElaOpt 
& n/A 
& n/A 
& 2006.8\\% & n/A\\

\bottomrule
\end{tabular}

\label{tab:computational_cost}
\end{table}

\subsection{Comparison of Proposed Refinement Frameworks}
\label{ssec:results_comparison}
We evaluate the proposed refinement frameworks on a large collection of AMASS sub-datasets using the SMPL target representation. The average across AMASS and several sub-dataset results are reported in Table~\ref{tab:comparison_all_methods} in terms of vertex-to-vertex error (V2V) and asymmetric Chamfer Distance (CD), both expressed in centimeters, together with the relative improvement over the ArtEq baseline. The remaining sub-datasets are found in Tables~\ref{tab:amass_subdatasets_supp_MVE}, \ref{tab:amass_subdatasets_supp_ACham} of appendix.

% \begin{table}[]
% \centering
% % \small
% \setlength{\tabcolsep}{0.5pt}
% % \renewcommand{\arraystretch}{.75}
% \caption{Comparison of reconstruction performance on representative AMASS sub-datasets. We report the mean vertex error (MVE) and Asymmetric Chamfer Distance (CD), both expressed in centimeters (cm), together with the relative improvement with respect to ArtEq. The column labeled AMASS does not include the BMLmovi sub-dataset as that was used as the training set for the two VaRefine methods.}
\begin{table}[t]
\centering
\setlength{\tabcolsep}{2pt}
\caption{Comparison of reconstruction performance on representative AMASS sub-datasets. ArtEq$^{+}$ denotes ArtEq retrained with the additional BMLmovi training frames. We report the vertex-to-vertex error (V2V) and Asymmetric Chamfer Distance (CD), both expressed in centimeters (cm), together with the relative improvement with respect to ArtEq. The column labeled AMASS does not include the BMLmovi sub-dataset as that was used as the training set for the two VaRefine methods.}
\resizebox{\columnwidth}{!}{%
\begin{tabular}{l|l|c|ccccc}
\toprule

\textbf{Method} 
& \textbf{Metric} 
& \textbf{AMASS} 
& \textbf{GRAB} 
& \textbf{TCD} 
& \textbf{HumanEva} 
& \textbf{EKUT} 
& \textbf{SFU} \\

\midrule

\multirow[c]{2}{*}{ArtEq}

& V2V
& \cellcolor{gray!8} 3.168
& \cellcolor{gray!8} 3.163
& \cellcolor{gray!8} 5.637
& \cellcolor{gray!8} 2.566
& \cellcolor{gray!8} 1.938
& \cellcolor{gray!8} 4.188 \\

& CD
& \cellcolor{gray!8} 1.681
& \cellcolor{gray!8} 1.708
& \cellcolor{gray!8} 2.689
& \cellcolor{gray!8} 1.327
& \cellcolor{gray!8} 1.087
& \cellcolor{gray!8} 2.223 \\

\midrule

\multirow[c]{2}{*}{ArtEq$^{+}$}

& V2V
& \cellcolor{gray!8} 2.401
& \cellcolor{gray!8} 1.994
& \cellcolor{gray!8} 3.092
& \cellcolor{gray!8} 2.257
& \cellcolor{gray!8} 2.043
& \cellcolor{gray!8} 3.031 \\

& CD
& \cellcolor{gray!8} 1.376
& \cellcolor{gray!8} 1.190
& \cellcolor{gray!8} 1.655
& \cellcolor{gray!8} 1.294
& \cellcolor{gray!8} 1.162
& \cellcolor{gray!8} 1.748 \\

\midrule
\multirow[c]{4}{*}{Latent-VaRefine}

& V2V
& \cellcolor{blue!8} 2.511
& \cellcolor{blue!8} 2.499
& \cellcolor{blue!8} 3.620
& \cellcolor{blue!8} 2.305
& \cellcolor{blue!8} 1.678
& \cellcolor{blue!8} 3.356\\

& Gain (\%)
& \cellcolor{blue!8} 20.7 
& \cellcolor{blue!8} 21.0
& \cellcolor{blue!8} 35.8
& \cellcolor{blue!8} 10.2
& \cellcolor{blue!8} 13.4 
& \cellcolor{blue!8} 19.9\\

\cmidrule(lr){2-8}

& CD
& \cellcolor{orange!8} 1.414
& \cellcolor{orange!8} 1.407
& \cellcolor{orange!8} 1.762
& \cellcolor{orange!8} 1.252
& \cellcolor{orange!8} 1.013
& \cellcolor{orange!8} 1.899 \\

& Gain (\%)
& \cellcolor{orange!8} 15.9
& \cellcolor{orange!8} 17.6
& \cellcolor{orange!8} 34.5
& \cellcolor{orange!8} 5.7
& \cellcolor{orange!8} 6.8
& \cellcolor{orange!8} 14.6 \\
\midrule

\multirow[c]{4}{*}{Vertex-VaRefine}

& V2V
& \cellcolor{blue!8} \textbf{1.552}
& \cellcolor{blue!8} \textbf{1.356}
& \cellcolor{blue!8} \textbf{2.457}
& \cellcolor{blue!8} \textbf{1.423}
& \cellcolor{blue!8} \textbf{1.094}
& \cellcolor{blue!8} 2.012 \\

& Gain (\%)
& \cellcolor{blue!8} \textbf{51.0}
& \cellcolor{blue!8} \textbf{57.1}
& \cellcolor{blue!8} \textbf{56.4}
& \cellcolor{blue!8} \textbf{44.5}
& \cellcolor{blue!8} \textbf{43.6}
& \cellcolor{blue!8} 52.0\\

\cmidrule(lr){2-8}

& CD
& \cellcolor{orange!8} 0.943

& \cellcolor{orange!8} 0.870
& \cellcolor{orange!8} 1.300
& \cellcolor{orange!8} 0.856
& \cellcolor{orange!8} 0.746
& \cellcolor{orange!8} 1.211 \\

& Gain (\%)
& \cellcolor{orange!8} 43.9
& \cellcolor{orange!8} 49.1
& \cellcolor{orange!8} 51.7
& \cellcolor{orange!8} 35.5
& \cellcolor{orange!8} 31.4
& \cellcolor{orange!8} 45.5\\

\midrule

\multirow[c]{4}{*}{ElaOpt}

& V2V
& \cellcolor{blue!8} 2.652
& \cellcolor{blue!8} 2.596
& \cellcolor{blue!8} 4.316
& \cellcolor{blue!8} 1.927
& \cellcolor{blue!8} 1.634
& \cellcolor{blue!8} 3.330\\

& Gain (\%)
& \cellcolor{blue!8} 16.3
& \cellcolor{blue!8} 17.9
& \cellcolor{blue!8} 23.4
& \cellcolor{blue!8} 24.9
& \cellcolor{blue!8} 15.7
& \cellcolor{blue!8} 20.5\\

\cmidrule(lr){2-8}

& CD
& \cellcolor{orange!8} 0.793
& \cellcolor{orange!8} 0.787
& \cellcolor{orange!8} 1.072
& \cellcolor{orange!8} 0.663
& \cellcolor{orange!8} 0.622
& \cellcolor{orange!8} 0.885 \\

& Gain (\%)
& \cellcolor{orange!8} 52.8
& \cellcolor{orange!8} 53.9
& \cellcolor{orange!8} 60.1
& \cellcolor{orange!8} 50.0
& \cellcolor{orange!8} 42.8
& \cellcolor{orange!8} 60.2 \\
\midrule

\multirow[c]{4}{*}{\shortstack{Latent-VaRefine \\+ ElaOpt}}

& V2V
& \cellcolor{blue!8} 2.008
& \cellcolor{blue!8} 1.979 
& \cellcolor{blue!8} 2.576 
& \cellcolor{blue!8} 1.673 
& \cellcolor{blue!8} 1.331 
& \cellcolor{blue!8} \textbf{1.943}  \\

& Gain (\%)
& \cellcolor{blue!8} 36.6
& \cellcolor{blue!8} 37.4
& \cellcolor{blue!8} 54.3
& \cellcolor{blue!8} 34.8
& \cellcolor{blue!8} 31.3
& \cellcolor{blue!8} \textbf{53.6}\\

\cmidrule(lr){2-8}

& CD
& \cellcolor{orange!8} \textbf{0.717} 
& \cellcolor{orange!8} \textbf{0.710} 
& \cellcolor{orange!8} \textbf{0.805} 
& \cellcolor{orange!8} \textbf{0.652} 
& \cellcolor{orange!8} \textbf{0.604 }
& \cellcolor{orange!8} \textbf{0.743}  \\

& Gain (\%)
& \cellcolor{orange!8} \textbf{57.3}
& \cellcolor{orange!8} \textbf{58.4}
& \cellcolor{orange!8} \textbf{70.1}
& \cellcolor{orange!8} \textbf{50.9}
& \cellcolor{orange!8} \textbf{44.4 }
& \cellcolor{orange!8} \textbf{66.6} \\

\midrule

\multirow[c]{4}{*}{\shortstack{Latent-VaRefine\\+ Vertex-VaRefine}}

& V2V
& \cellcolor{blue!8} 1.715

& \cellcolor{blue!8} 1.558
& \cellcolor{blue!8} 2.529
& \cellcolor{blue!8} 1.593
& \cellcolor{blue!8} 1.264
& \cellcolor{blue!8} 2.599 \\

& Gain (\%)
& \cellcolor{blue!8} 45.9
& \cellcolor{blue!8} 50.7
& \cellcolor{blue!8} 55.1
& \cellcolor{blue!8} 37.9
& \cellcolor{blue!8} 34.8
& \cellcolor{blue!8} 37.9\\

\cmidrule(lr){2-8}

& CD
& \cellcolor{orange!8} 1.010

& \cellcolor{orange!8} 0.946 
& \cellcolor{orange!8}  1.317
& \cellcolor{orange!8}  0.923
& \cellcolor{orange!8}  0.819
& \cellcolor{orange!8}  1.497\\

& Gain (\%)
& \cellcolor{orange!8} 39.9
& \cellcolor{orange!8} 44.6
& \cellcolor{orange!8} 51.0
& \cellcolor{orange!8} 30.4
& \cellcolor{orange!8} 24.7
& \cellcolor{orange!8}  32.7\\

\bottomrule
\end{tabular}%
}
\label{tab:comparison_all_methods}
\end{table}

% \label{tab:comparison_all_methods}

% \end{table}

\begin{figure}[]
    \centering
    \includegraphics[width=\linewidth]{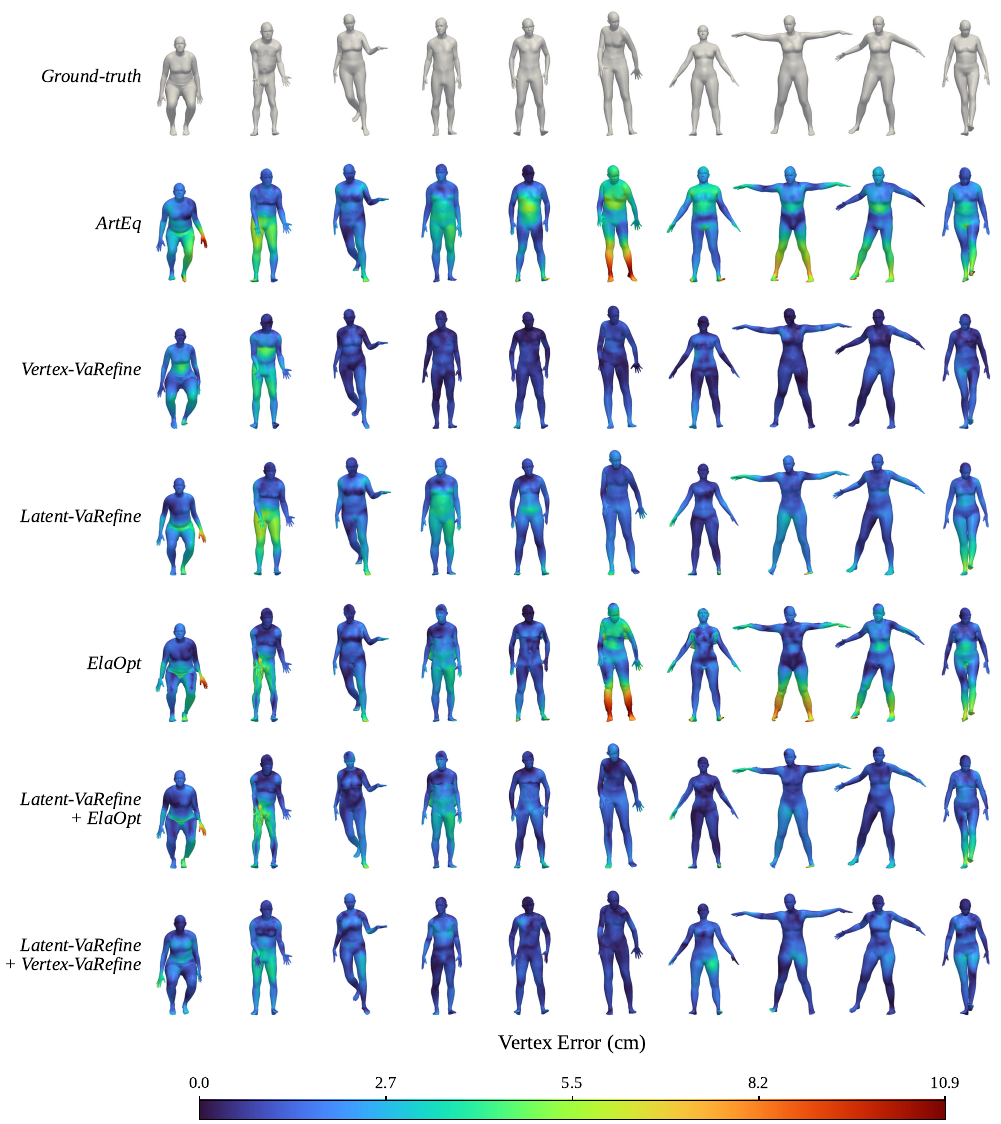}
    \caption{Qualitative comparison between ArtEq, latent-VaRefine, Vertex-VaRefine, ElaOpt, Latent-VaRefine + ElaOpt, and Latent-VaRefine + Vertex-VaRefine on different body shapes and poses. Color maps show the per-vertex reconstruction error with respect to the ground-truth surface in centimeters (cm).}
    \label{fig:Heatmap_results}
\end{figure}

Overall, all refinement strategies improve the reconstruction quality across all evaluated datasets, demonstrating strong generalization capabilities across diverse motion distributions and capture conditions. Among the learning-based approaches, Vertex-VaRefine consistently achieves the lowest V2V values. In particular, large improvements are observed on challenging datasets such as GRAB, TCD, and SFU, where Vertex-VaRefine reduces the V2V by up to $57.1\%$, $56.4\%$, and $52.0\%$, respectively.
The optimization based method, ElaOpt, achieves the best Chamfer Distance results across most datasets, with improvements exceeding $50\%$ on several benchmarks. Latent-VaRefine already provides a noticeable improvement over the initial ArtEq reconstructions, but remains less accurate than Vertex-VaRefine on both V2V and CD metrics, and than ElaOpt on CD. Its main advantage, however, is that it stays within the SMPL latent space: it outputs refined SMPL parameters rather than a per-vertex displacements, thereby preserving their interpretability. These observations confirm that directly refining the reconstructed geometry at the vertex level leads to more accurate and stable surface corrections than operating only in the latent space, at the cost of leaving the original parametric representation. Finally, we also report results of combining Latent-VaRefine with ElaOpt, by running these two frameworks in succession, which turns out to lead to the consistently best CD results across all data sets and both comparison methods. We also retrain ArtEq on the additional data used in our pipeline (i.e., on DFAUST+BMLmovi), indicated as ArtEq$^{+}$. Although this does reduce ArtEq's error, it clearly remains behind Vertex-VaRefine, Latent-VaRefine + ElaOpt, Latent-VaRefine + Vertex-VaRefine applied on top of the original ArtEq, on all sub-datasets, corroborating the fact that the \emph{improvement is not explained by the additional training data alone}. In addition to these quantitative results, we present in Figure~\ref{fig:Heatmap_results} a  qualitative comparison, which  demonstrates that all proposed methods effectively reduce the per-vertex reconstruction errors across a wide variety of body shapes and poses compared to the initial ArtEq reconstructions.

\subsection{Evaluation with Additional Metrics}
In order to test the independence of our results to the choice of metric, we report the results on one subset of AMASS using an additional set of metrics that are independent of the training objectives. For the SSM\_synced subset of AMASS, Table~\ref{tab:additional_metrics} reports the relative improvement over ArtEq under Hausdorff distance, vertex-to-vertex supremum, and both oriented and unoriented varifold metrics (using \(\sigma=.1\)). Interestingly, the improvements under these metrics are of the same order as those obtained with the V2V and Chamfer distances used in the design and training of the methods. The descriptions of these distances can be found in the Appendix \ref{sec:distances}.

\begin{table}[]
\centering
\small
\setlength{\tabcolsep}{2pt}
\renewcommand{\arraystretch}{1}
\caption{Relative improvement (\%) over the ArtEq baseline on the SSM\_synced
subset of AMASS, under metrics independent of the training objective.}
\resizebox{\columnwidth}{!}{%

\begin{tabular}{lccc}
\toprule
\textbf{Metric}
& \textbf{Latent-VaRefine}
& \textbf{Vertex-VaRefine}
& \textbf{ElaOpt} \\
\midrule
Hausdorff              & 43.5 & 57.1  & \textbf{60.1} \\
V2V supremum           & 41.3 & \textbf{55.9} & 18.3 \\
Varifold (oriented)    & 32.7 & 59.0 & \textbf{75.4} \\
Varifold (unoriented)  & 32.7 & 57.2  & \textbf{75.2} \\
\bottomrule
\end{tabular}%
}
\label{tab:additional_metrics}
\end{table}

\section{Ablation Studies}
\subsection{Number of VariGrad Scales} We analyze here the contribution of the proposed components through a series of ablation studies. First, we evaluate the impact of the number of VariGrad scales used during refinement. We study the effect of different multi-scale VariGrad configurations on the performance of the Latent-VaRefine and Vertex-VaRefine models. In the latent refinement setting, increasing the number of scales beyond a single-scale representation slightly degrades performance. In contrast, the vertex refinement model benefits from multi-scale VariGrad features with the 3-scale configuration providing the best trade-off between accuracy and stability across the AMASS benchmark and is therefore adopted in the final model.
%\todomartin{Describe findings and refer to supp material!}

\paragraph{Latent-VaRefine} For Latent-VaRefine, we consider 3 and 5 scale variants and also perform an ablation to test the value of including the coefficient fidelity term. In addition to the base \(.75\) cm base scale, we include half and double scale (\(.375\) cm and \(1.5\) cm) VariGrad descriptors to obtain a 3-scale VariGrad layer. To these are further added the \(5\) cm and \(15\) cm VariGrads to obtain a 5-scale model. The ablation to test the value of the coefficient fidelity term is performed using the same single \(.75\) cm VariGrad as the standard Latent-VaRefine with the only difference being that only MSE is used for training loss in this case. 

The comparison of the standard 1 scale Latent-VaRefine to the 3 and 5 scale versions is found in Table~\ref{tab:ablation_scales}, which also includes the MSE-only ablation results. In all cases, the standard Latent-VaRefine performs best by a slim margin. It is likely that the wider model needed to accommodate the additional input data is the cause of the lower performance of the 3 and 5 scale variants. The inclusion of the coefficient fidelity term appears to improve both MVE and Chamfer error as the term likely both regularizes the training and directly encourages improvement as errors in coefficients lead to higher MVE and Chamfer error.
%In the first ablation study we perform on Latent-VarRefine, we experiment with using multiple scales for the VariGrad representation.  To these are further added the \(5\) cm and \(15\) cm VariGrads resulting in a 5-scale model. We found numerically that both of these multiscale VariGrads performed slightly worse than the single-scale version on this particular model, likely due to the widening and increase in trainable parameters needed to account for a larger input.
%
%The second ablation study performed on Latent-VarRefine justifies the necessity of the coefficient fidelity term in the loss function. Removing this term in training tends to slightly worsen the model's results, as shown in table~\ref{tab:ablation_scales}. Coefficient loss likely both regularizes learning and directly encourages it, as errors in coefficients will lead to higher error in both MVE and Chamfer measurements.  

\paragraph{Vertex-VaRefine}
Similar to the study on Latent-VaRefine, we wish to determine the potential effect of adding more scales in the VariGrad representation. To that end, we consider multiple uniformly spaced scales in order to capture geometric discrepancies at different spatial resolutions. We thus consider three specific configurations: a single scale \{5 cm\}, three scales \{3 cm, 6 cm, 10 cm\}, and five scales \{2 cm, 4 cm, 6 cm, 8 cm, 10 cm\}. Table~\ref{tab:ablation_scales} reports the reconstruction performance obtained with each configuration using both the vertex-to-vertex (V2V) and the asymmetric Chamfer Distance (CD). These results show that multi-scale VariGrad features consistently improve reconstruction performance over a single scale. Notably, the performance gain between the single-scale and the 3-scale configuration (V2V: 1.734 vs. 1.552 on AMASS) is substantially larger than the performance gain between the 3-scale and 5-scale configurations (V2V: 1.552 vs. 1.527), indicating that most of the benefit of multi-scale VariGrad features is already achieved with 3 scales, with diminishing returns beyond that point. While the 5-scale configuration achieves the best results on most datasets, this additional improvement over the 3-scale configuration is modest and comes at a noticeably higher memory and training-time cost, since each additional scale requires computing, storing, and projecting an extra VariGrad feature channel per vertex. We therefore adopt the 3-scale configuration as our final choice, as it offers the best trade-off between reconstruction accuracy and computational cost.

\definecolor{latentcolor}{RGB}{222, 235, 250}
\definecolor{vertexcolor}{RGB}{226, 245, 221}

\begin{table}
\centering
\small
\setlength{\tabcolsep}{3pt}
\renewcommand{\arraystretch}{1.1}
\caption{Ablation studies on the number of VariGrad scales for Latent-VaRefine
and Vertex-VaRefine, and on the coefficient loss term of Latent-VaRefine. We
report vertex-to-vertex (V2V) and asymmetric Chamfer Distance (CD) errors,
both in cm, on representative datasets and on the entire AMASS benchmark
(excluding BMLmovi). Best results are in bold, second-best are underlined.}
\resizebox{\columnwidth}{!}{%
\begin{tabular}{l|l|l|c|ccc}
\toprule
\textbf{Model} & \textbf{Configuration} & \textbf{Metric} & \textbf{AMASS} & \textbf{GRAB} & \textbf{EKUT} & \textbf{SFU} \\
\midrule
\multicolumn{2}{l|}{\multirow{2}{*}{ArtEq}}
 & V2V & \cellcolor{gray!8} 3.168 & \cellcolor{gray!8} 3.163 & \cellcolor{gray!8} 1.938 & \cellcolor{gray!8} 4.188 \\
\multicolumn{2}{l|}{}
 & CD  & \cellcolor{gray!8} 1.681 & \cellcolor{gray!8} 1.708 & \cellcolor{gray!8} 1.087 & \cellcolor{gray!8} 2.223 \\
\midrule
\multirow{8}{*}{\shortstack[l]{Latent-\\VaRefine}}
 & \multirow{2}{*}{1 scale}
   & V2V & \cellcolor{latentcolor} \textbf{2.511} & \cellcolor{latentcolor} \textbf{2.499} & \cellcolor{latentcolor} \textbf{1.678} & \cellcolor{latentcolor} \textbf{3.356} \\
 & & CD  & \cellcolor{latentcolor} \textbf{1.414} & \cellcolor{latentcolor} \textbf{1.407} & \cellcolor{latentcolor} \textbf{1.013} & \cellcolor{latentcolor} \textbf{1.899} \\
\cmidrule(l){2-7}
 & \multirow{2}{*}{\shortstack[l]{1 scale\\(MSE loss only)}}
   & V2V & \cellcolor{latentcolor} 2.625 & \cellcolor{latentcolor} 2.577 & \cellcolor{latentcolor} 1.722 & \cellcolor{latentcolor} 3.525 \\
 & & CD  & \cellcolor{latentcolor} 1.470 & \cellcolor{latentcolor} 1.466 & \cellcolor{latentcolor} 1.024 & \cellcolor{latentcolor} 1.965 \\
\cmidrule(l){2-7}
 & \multirow{2}{*}{3 scales}
   & V2V & \cellcolor{latentcolor} \underline{2.545} & \cellcolor{latentcolor} \underline{2.528} & \cellcolor{latentcolor} \underline{1.703} & \cellcolor{latentcolor} 3.431 \\
 & & CD  & \cellcolor{latentcolor} \underline{1.433} & \cellcolor{latentcolor} 1.425 & \cellcolor{latentcolor} \underline{1.021} & \cellcolor{latentcolor} 1.934 \\
\cmidrule(l){2-7}
 & \multirow{2}{*}{5 scales}
   & V2V & \cellcolor{latentcolor} 2.590 & \cellcolor{latentcolor} 2.529 & \cellcolor{latentcolor} 1.721 & \cellcolor{latentcolor} \underline{3.394} \\
 & & CD  & \cellcolor{latentcolor} 1.440 & \cellcolor{latentcolor} \underline{1.417} & \cellcolor{latentcolor} 1.022 & \cellcolor{latentcolor} \underline{1.918} \\
\midrule
\multirow{6}{*}{\shortstack[l]{Vertex-\\VaRefine}}
 & \multirow{2}{*}{1 scale}
   & V2V & \cellcolor{vertexcolor} 1.734 & \cellcolor{vertexcolor} 1.466 & \cellcolor{vertexcolor} 1.153 & \cellcolor{vertexcolor} 2.268 \\
 & & CD  & \cellcolor{vertexcolor} 1.043 & \cellcolor{vertexcolor} 0.950 & \cellcolor{vertexcolor} 0.788  & \cellcolor{vertexcolor} 1.338 \\
\cmidrule(l){2-7}
 & \multirow{2}{*}{3 scales}
   & V2V & \cellcolor{vertexcolor} \underline{1.552} & \cellcolor{vertexcolor} \underline{1.356} & \cellcolor{vertexcolor} \underline{1.094} & \cellcolor{vertexcolor} \underline{2.012} \\
 & & CD  & \cellcolor{vertexcolor} \underline{0.943} & \cellcolor{vertexcolor} \underline{0.870} & \cellcolor{vertexcolor} \underline{0.746} & \cellcolor{vertexcolor} \textbf{1.211} \\
\cmidrule(l){2-7}
 & \multirow{2}{*}{5 scales}
   & V2V & \cellcolor{vertexcolor} \textbf{1.527} & \cellcolor{vertexcolor} \textbf{1.280} & \cellcolor{vertexcolor} \textbf{0.964} & \cellcolor{vertexcolor} \textbf{1.959} \\
 & & CD  & \cellcolor{vertexcolor} \textbf{0.933} & \cellcolor{vertexcolor} \textbf{0.828} & \cellcolor{vertexcolor} \textbf{0.700} & \cellcolor{vertexcolor} \underline{1.226} \\
\bottomrule
\end{tabular}%
}
\label{tab:ablation_scales}
\end{table}

\subsection{Effect of the Equivariance in Vertex-VaRefine}
\label{sec:ablation_equivariance}
Section~\ref{ssec:vertex_varefine} argues that Vertex-VaRefine is approximately pose-equivariant \emph{by construction}, but not fully equivariant; this is confirmed by Table~\ref{tab:varigrad_pose_approx}, with a median relative error of $4$--$10\%$ and a mean relative error of $23$--$50\%$. To determine if this approximate equivariance is genuinely beneficial rather than merely a design choice, we trained an otherwise identical variant that entirely omits the canonical-frame projection and the neutral pose/re-posing step. In this version, the DiffusionNet backbone processes raw, unprojected VariGrad descriptors computed directly on the posed source mesh and predicts displacement directly in posed space, providing no equivariance guarantees. Both variants maintain the same backbone architecture, VariGrad scales, loss functions, and training data; the only difference lies in the pose-equivariant wrapping.

\begin{table}[t]
\centering
\setlength{\tabcolsep}{2pt}
\caption{Ablation of the pose-equivariant construction in Vertex-VaRefine (DiffusionNet backbone), evaluated on the same representative AMASS sub-datasets as Table~\ref{tab:comparison_all_methods}. Both variants share identical backbone, VariGrad features, training data, and schedule; the only difference is the canonical T-pose projection and re-posing step of Section~\ref{ssec:vertex_varefine}. We report the vertex-to-vertex error (V2V) and Asymmetric Chamfer Distance (CD), in centimeters (cm), together with the relative gain (\%) over the ArtEq baseline. The AMASS column is the average over the AMASS sub-datasets, excluding BMLmovi, which was used for training. Removing the equivariant construction degrades both metrics on every sub-dataset.}

\resizebox{\columnwidth}{!}{%
\begin{tabular}{l|l|c|ccccc}
\toprule

\textbf{Method} 
& \textbf{Metric} 
& \textbf{AMASS} 
& \textbf{GRAB} 
& \textbf{TCD} 
& \textbf{HumanEva} 
& \textbf{EKUT} 
& \textbf{SFU} \\

\midrule

\multirow[c]{2}{*}{ArtEq}

& V2V
& \cellcolor{gray!8} 3.168
& \cellcolor{gray!8} 3.163
& \cellcolor{gray!8} 5.637
& \cellcolor{gray!8} 2.566
& \cellcolor{gray!8} 1.938
& \cellcolor{gray!8} 4.188 \\

& CD
& \cellcolor{gray!8} 1.681
& \cellcolor{gray!8} 1.708
& \cellcolor{gray!8} 2.689
& \cellcolor{gray!8} 1.327
& \cellcolor{gray!8} 1.087
& \cellcolor{gray!8} 2.223 \\

\midrule

\multirow[c]{4}{*}{\shortstack{Vertex-VaRefine\\(w/ equiv.)}}

& V2V
& \cellcolor{blue!8} \textbf{1.552}
& \cellcolor{blue!8} \textbf{1.356}
& \cellcolor{blue!8} \textbf{2.457}
& \cellcolor{blue!8} \textbf{1.423}
& \cellcolor{blue!8} \textbf{1.094}
& \cellcolor{blue!8} \textbf{2.012} \\

& Gain (\%)
& \cellcolor{blue!8} \textbf{51.0}
& \cellcolor{blue!8} \textbf{57.1}
& \cellcolor{blue!8} \textbf{56.4}
& \cellcolor{blue!8} \textbf{44.5}
& \cellcolor{blue!8} \textbf{43.6}
& \cellcolor{blue!8} \textbf{52.0} \\

\cmidrule(lr){2-8}

& CD
& \cellcolor{orange!8} \textbf{0.943}
& \cellcolor{orange!8} \textbf{0.870}
& \cellcolor{orange!8} \textbf{1.300}
& \cellcolor{orange!8} \textbf{0.856}
& \cellcolor{orange!8} \textbf{0.746}
& \cellcolor{orange!8} \textbf{1.211} \\

& Gain (\%)
& \cellcolor{orange!8} \textbf{43.9}
& \cellcolor{orange!8} \textbf{49.1}
& \cellcolor{orange!8} \textbf{51.7}
& \cellcolor{orange!8} \textbf{35.5}
& \cellcolor{orange!8} \textbf{31.4}
& \cellcolor{orange!8} \textbf{45.5} \\

\midrule

\multirow[c]{4}{*}{\shortstack{Vertex-VaRefine\\(w/o equiv.)}}

& V2V
& \cellcolor{blue!8} 1.666
& \cellcolor{blue!8} 1.554
& \cellcolor{blue!8} 2.670
& \cellcolor{blue!8} 1.607
& \cellcolor{blue!8} 1.192
& \cellcolor{blue!8} 2.540 \\

& Gain (\%)
& \cellcolor{blue!8} 47.4
& \cellcolor{blue!8} 50.9
& \cellcolor{blue!8} 52.6
& \cellcolor{blue!8} 37.4
& \cellcolor{blue!8} 38.5
& \cellcolor{blue!8} 39.4 \\

\cmidrule(lr){2-8}

& CD
& \cellcolor{orange!8} 1.001
& \cellcolor{orange!8} 0.975
& \cellcolor{orange!8} 1.385
& \cellcolor{orange!8} 0.941
& \cellcolor{orange!8} 0.782
& \cellcolor{orange!8} 1.455 \\

& Gain (\%)
& \cellcolor{orange!8} 40.5
& \cellcolor{orange!8} 42.9
& \cellcolor{orange!8} 48.5
& \cellcolor{orange!8} 29.1
& \cellcolor{orange!8} 28.1
& \cellcolor{orange!8} 34.5 \\

\bottomrule
\end{tabular}%
}
\label{tab:diffusionnet_equiv_ablation}
\end{table}

Table~\ref{tab:diffusionnet_equiv_ablation} demonstrates that omitting this construction impairs both metrics across all sub-datasets. On AMASS, the V2V gain decreases from $51\%$ to $47.4\%$, and the CD gain from $43.9\%$ to $40.5\%$. The decline is most pronounced on SFU, where the V2V gain falls from $52.0\%$ to $39.4\%$. Consequently, despite the input features being only approximately pose-invariant, the proposed construction yields a consistent performance improvement.

\subsection{Backbone Choice: DiffusionNet vs.\ SpiralNet vs.\ PoissonNet}
\label{sec:ablation_backbones}
% \todonicolas{Where are the results of this ablation study? I don't see it anywhere in this section or in Appendix C.}\textcolor{blue}{Do you mean qualitative results? Because the quantitative results are presented in Table 7.}
Vertex-VaRefine's pipeline -- VariGrad computation, projection onto
per-vertex rest-pose frames (Eq.~\eqref{eq:varigrad_frame_projection}),
T-pose prediction, and re-posing by linear blend skinning
(Eq.~\eqref{eq:tpose_repose}) -- is agnostic to the network processing
$\phi$. We assess this choice by replacing the DiffusionNet backbone
(Section~\ref{ssec:vertex_varefine}) with two alternatives, both are described in Appendix~\ref{app:backbones}, trained and
evaluated identically as in 
Section~\ref{sec:implementation_details}. \textbf{PoissonNet} uses a
PoissonNet~\citep{maesumi2025poissonnet} backbone of the same width and
depth as DiffusionNet ($C_{\text{width}}{=}128$, $N_{\text{block}}{=}4$),
replacing learned diffusion with a per-block Poisson solve -- involving one Cholesky
factorization of the cotangent Laplacian per mesh~\citep{nicolet2021large}
-- followed by the same linear displacement head. We also test the \textbf{SpiralNet} approach with a SpiralNet++~\citep{gong2019spiralnet++} encoder
(three spiral convolution layers, $\{32,32,64\}$, ReLU, no downsampling),
whose per-vertex output is averaged per face into $64$-dim descriptors and
fed to a \textbf{Neural Jacobian
Field}~\citep{NeuralJacobianFieldACMTG2022} decoder (4 FC layers, $128$
units, GroupNorm, ReLU) that predicts a per-face $3{\times}3$ Jacobian.
Unlike the other two backbones, the displacement is not a direct output:
the Jacobian field is integrated via a Poisson solve factorized once on a
fixed template $\mathcal{S}_0$, and the resulting correction $\Delta$ is
added to the per-sample neutral pose shape, $D\doteq\Delta$,
$\restpose=\bar{\mathcal{S}}+\Delta$ -- keeping this head compatible with the
shared equivariant pipeline. We report the results in Table~\ref{tab:ablation_backbones}, confirming DiffusionNet as the optimal choice for this framework.

\begin{table}[H]
\centering
\small
\setlength{\tabcolsep}{1.5pt}
\renewcommand{\arraystretch}{1}
\caption{Ablation on the backbone processing the frame-projected VariGrad
features $\phi$. All three variants share the same descriptors, rest-pose
frame projection, T-pose prediction, re-posing, loss and training schedule;
only the backbone (and, for SpiralNet++, the decoding head it requires)
changes. V2V and asymmetric Chamfer Distance in centimeters.}
\resizebox{\columnwidth}{!}{%

\begin{tabular}{llcccc}
\toprule

\textbf{Subset} & \textbf{Metric}
& \textbf{ArtEq}
& \makecell{\textbf{DiffusionNet}\\\textbf{(Vertex-VaRefine)}}
& \textbf{SpiralNet}
& \textbf{PoissonNet} \\

\midrule

\multirow{2}{*}{AMASS}
& V2V & 3.168 & \textbf{1.552}  & 2.019  & 2.118  \\
& CD  & 1.681 & \textbf{0.943} & 1.162 & 1.205  \\

\midrule

\multirow{2}{*}{SFU}
& V2V & 4.188 & \textbf{2.012} & 2.612  & 2.823  \\
& CD  & 2.223 & \textbf{1.211}  & 1.486  & 1.592  \\

\bottomrule
\end{tabular}%
}
\label{tab:ablation_backbones}
\end{table}

\subsection{ElaOpt Ablations} 

We perform two ablation studies on ElaOpt. The first is a simple comparison against \emph{SMPL+D}, and the latter is a term-wise examination, to qualitatively inspect the value of the individual terms and their quantitative impact on reported metrics such as V2V error and Chamfer error.

\paragraph{Comparison With \emph{SMPL+D}} \label{smplDandElaOpt}The results of an unrestricted SMPL+D-style fitting are less accurate and regular than those of ElaOpt; for example, on the SFU dataset a SMPL+D style fitting with no regularizer and an SRNF regularizer produces a CD improvement of 24\%, 31\% respectively.

\paragraph{Termwise Ablation} \label{termwiseElaOptTest}For the second ablation, we consider six different variants of ElaOpt. The first is ElaOpt as presented in \autoref{ELaOptInternals}. The second is plain, normal-restrained optimization. To this are added masking and the \(H^1\), \(H^2\), and SRNF terms. All six cases retain the constraint to the normal direction, differentiating them from an entirely free optimization. 

\begin{figure*}
    \centering
    \includegraphics[width=1\linewidth]{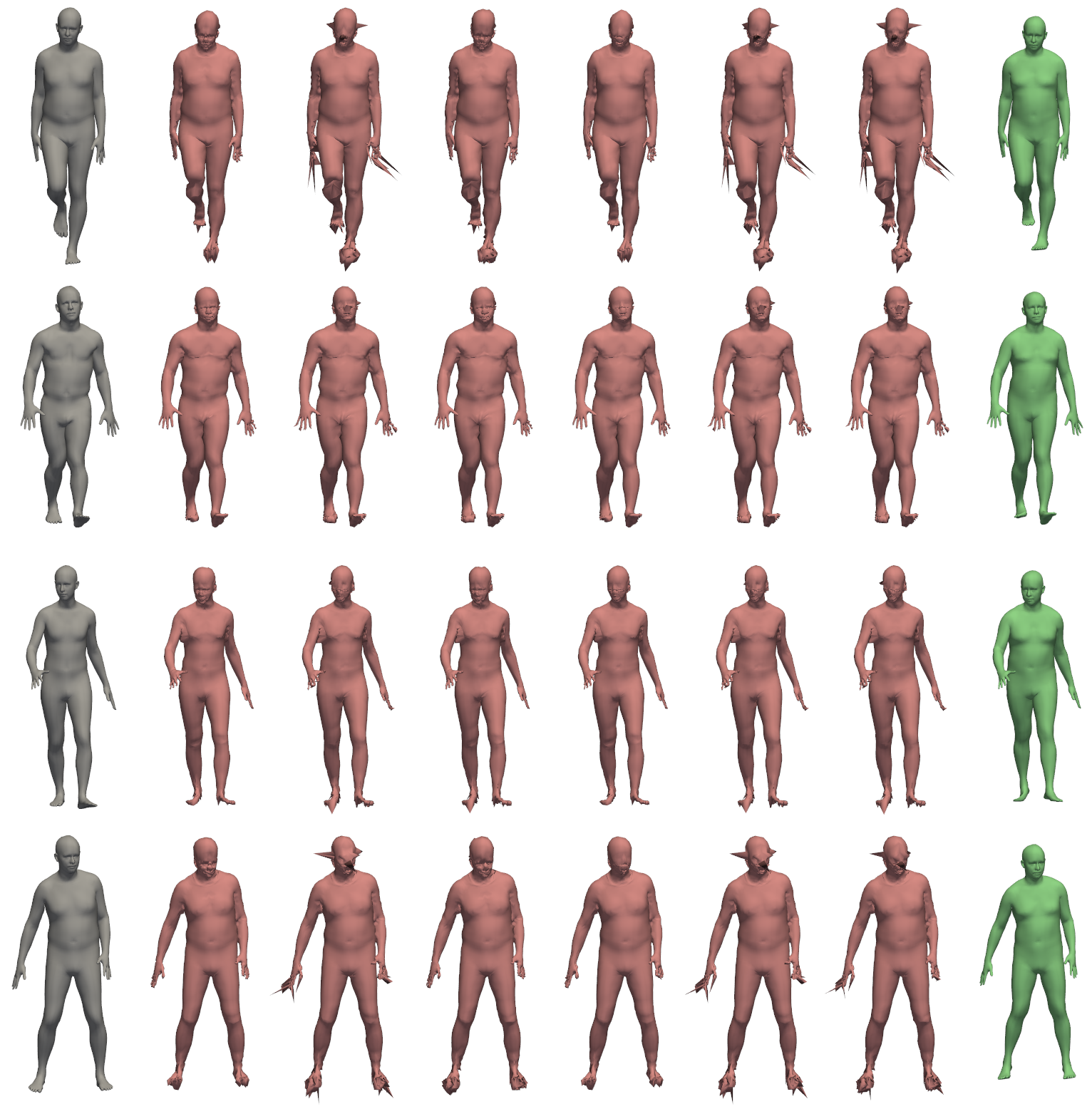}
\caption{Comparison of the methods used in the ElaOpt term-wise ablation on four meshes. On the left in gray is the target to be optimized to and on the right, in green, is the base ArtEq mesh each method will be applied to. In red, from left to right are: ElaOpt, plain normal-constrained optimization, normal-constrained optimization with masks, with \(H^1\), with \(H^2\), and with \(\mathrm{SNRF}\). Note the numerical instability artifacts in the plain optimization as well as the distorted faces in those methods not using masks.}
    \label{fig:ElaOpt_termwisesample}
\end{figure*}

A sample of qualitative results can be seen in Figure \ref{fig:ElaOpt_termwisesample} and the quantitative results across the SFU dataset can be found in Table \ref{tab:ablation_Ela}. Due to the nature of optimization methods, it can be noted that while the quantitative results of less restrained methods can get lower numerically, this is not always preferable if it comes at the cost of a messier, qualitatively worse result.

Although quantitatively ElaOpt does not win in V2V nor Chamfer metrics, the qualitative value of the fully regularized method can be seen in Figure~\ref{fig:ElaOpt_termwisesample}. In this figure, we can see the value of the masks in retaining an identifiable face and the regularization terms for their avoidance of numerical instability which can be seen in the figure as jagged artifacts.

\begin{table}[H]
\centering
\small
\setlength{\tabcolsep}{1.5pt}
\renewcommand{\arraystretch}{1}
\caption{Numerical evaluation on the effect of the ElaOpt terms on V2V and Chamfer. All computations are on the SFU dataset and reported in cm.}
\resizebox{\columnwidth}{!}{%

\begin{tabular}{cccccccc}
\toprule

&\textbf{ArtEq} & \textbf{ElaOpt} & \textbf{Nothing}& \textbf{+Masks}
& \textbf{+\(H^1\)}
& \textbf{+\(H^2\)} & \textbf{+SRNF}  \\

\midrule

\textbf{V2V} & 4.188 & 3.330 & 3.270& 3.314 & 3.265 & 3.268 & 3.314 \\

\midrule

\textbf{CD} & 2.223 & 0.885 &0.848& 0.861& 0.850& 0.908 &0.851  \\

\bottomrule
\end{tabular}
}
\label{tab:ablation_Ela}
\end{table}

\section{Generalizability}
\label{sec:generalizability}

As a final set of evaluation tests, we examine the generalizability of our methods. We show that they work with a different choice of decoder (OmniFit~\citep{cai2026omnifit}), of latent space (SMPL-H/X), and with targets outside of the reach of the decoder (due to either a change of resolution or belonging to a more general category, such as raw scans). 

\subsection{Generalizability to a different latent space} To demonstrate that our framework is not specific to SMPL, we apply all three methods (Latent-VaRefine, Vertex-VaRefine and ElaOpt) to two richer body models. SMPL-H shares the SMPL template topology, but articulates the fingers. SMPL-X additionally models the jaw, eyes, and facial expressions. Since ArtEq is trained for SMPL, we do not retrain it for these models.
Instead, the SMPL-H and SMPL-X codes of both the source and target meshes are obtained from the corresponding SMPL codes with the SMPL$\to$SMPL-H and SMPL$\to$SMPL-X converters of Appendix~\ref{app:converter}: the source codes are converted ArtEq predictions, and the target codes are converted ground-truth SMPL codes. The two learning-based methods are retrained with the SMPL-H/SMPL-X versions of the BMLmovi data, keeping the architectures of Section~\ref{sec:implementation_details}. In the case of Vertex-VaRefine, hyper-parameters are also kept identical. For Latent-VaRefine the learning rate and relative coefficient-loss/MSE-loss weighting were adjusted. For SMPL-X we set \(w_{\mathrm{MSE}}=20\) while for SMPL-H we set \(w_{\mathrm{MSE}}=1\). For both, the initial learning rate was set to \(10^{-4.0}\) and an additional weight was needed for the new hand parameters for the latent space loss. As there are 90 hand parameters in SMPL-H and SMPL-X and these parameters have little effect on overall mesh correctness, this weight was set to \(0\).  ElaOpt requires no training and is applied directly to the converted source and target meshes. All methods are evaluated on the remaining AMASS sub-datasets, and are showed to perform similarly well as on their SMPL version.

\begin{table}[]
\centering
\small
\setlength{\tabcolsep}{1pt}
\caption{Extension to SMPL-H and SMPL-X (neutral models). V2V and asymmetric
Chamfer Distance (CD) in cm, with relative improvement (Gain) over the initial
source reconstruction. The AMASS column does not include BMLmovi, which is
used for training.}
\resizebox{\columnwidth}{!}{%
\begin{tabular}{l|l|ccc|ccc}
\toprule
& & \multicolumn{3}{c|}{\textbf{SMPL-H}} & \multicolumn{3}{c}{\textbf{SMPL-X}} \\
\textbf{Method} & \textbf{Metric} & AMASS & TCD & SFU & AMASS & TCD & SFU \\
\midrule

\multirow{2}{*}{ArtEq}
& V2V
& \cellcolor{gray!8} 3.167
& \cellcolor{gray!8} 5.637
& \cellcolor{gray!8} 4.188
& \cellcolor{gray!8} 2.575
& \cellcolor{gray!8} 4.282
& \cellcolor{gray!8} 3.805 \\
& CD
& \cellcolor{gray!8} 1.681
& \cellcolor{gray!8} 2.689
& \cellcolor{gray!8} 2.223
& \cellcolor{gray!8} 1.296
& \cellcolor{gray!8} 1.908
& \cellcolor{gray!8} 1.807 \\
\midrule

\multirow[c]{4}{*}{\shortstack{Latent-\\VaRefine}}
& V2V
& \cellcolor{blue!8} 2.675
& \cellcolor{blue!8} 4.103
& \cellcolor{blue!8} 3.600
& \cellcolor{blue!8} 
2.138 & \cellcolor{blue!8} 
2.923 & \cellcolor{blue!8}  3.072\\
& Gain (\%)
& \cellcolor{blue!8} 15.5 
& \cellcolor{blue!8} 27.2
& \cellcolor{blue!8} 
14.0
& \cellcolor{blue!8} 17.0
& \cellcolor{blue!8} 31.7
& \cellcolor{blue!8} 19.3  \\
\cmidrule(lr){2-8}
& CD
& \cellcolor{orange!8} 1.480
& \cellcolor{orange!8}
2.035
& \cellcolor{orange!8}
2.012
& \cellcolor{orange!8}1.119
& \cellcolor{orange!8} 1.361
& \cellcolor{orange!8}
1.546\\
& Gain (\%)
& \cellcolor{orange!8} 12.0
& \cellcolor{orange!8} 24.3
& \cellcolor{orange!8} 9.5
& \cellcolor{orange!8} 13.7
& \cellcolor{orange!8} 28.7
& \cellcolor{orange!8} 14.4 \\
\midrule

\multirow[c]{4}{*}{\shortstack{Vertex-\\VaRefine}}
& V2V
& \cellcolor{blue!8} \textbf{1.445}
& \cellcolor{blue!8} \textbf{2.605}
& \cellcolor{blue!8} \textbf{2.219}
& \cellcolor{blue!8} \textbf{1.188}
& \cellcolor{blue!8} \textbf{1.798}
& \cellcolor{blue!8} \textbf{1.849} \\
& Gain (\%)
& \cellcolor{blue!8} \textbf{54.4}
& \cellcolor{blue!8} \textbf{53.8}
& \cellcolor{blue!8} \textbf{47.0}
& \cellcolor{blue!8} \textbf{53.9}
& \cellcolor{blue!8} \textbf{58.0}
& \cellcolor{blue!8} \textbf{51.4} \\
\cmidrule(lr){2-8}
& CD
& \cellcolor{orange!8} 0.904
& \cellcolor{orange!8} 1.354
& \cellcolor{orange!8} 1.364
& \cellcolor{orange!8} 0.688
& \cellcolor{orange!8} \textbf{0.918}
& \cellcolor{orange!8} 1.042 \\
& Gain (\%)
& \cellcolor{orange!8} 46.2
& \cellcolor{orange!8} 49.6
& \cellcolor{orange!8} 38.6
& \cellcolor{orange!8} 46.9
& \cellcolor{orange!8} \textbf{51.9}
& \cellcolor{orange!8} 42.3 \\
\midrule

\multirow[c]{4}{*}{ElaOpt}
& V2V
& \cellcolor{blue!8} 2.616
& \cellcolor{blue!8} 4.426
& \cellcolor{blue!8} 3.442
& \cellcolor{blue!8} 2.063
& \cellcolor{blue!8} 4.097
& \cellcolor{blue!8} 2.998 \\
& Gain (\%)
& \cellcolor{blue!8} 17.4
& \cellcolor{blue!8} 21.5
& \cellcolor{blue!8} 17.8
& \cellcolor{blue!8} 19.9
& \cellcolor{blue!8} 4.3
& \cellcolor{blue!8} 21.2 \\
\cmidrule(lr){2-8}
& CD
& \cellcolor{orange!8} \textbf{0.803}
& \cellcolor{orange!8} \textbf{1.204}
& \cellcolor{orange!8} \textbf{0.978}
& \cellcolor{orange!8} \textbf{0.608}
& \cellcolor{orange!8} 0.997
& \cellcolor{orange!8} \textbf{0.831} \\
& Gain (\%)
& \cellcolor{orange!8} \textbf{52.2}
& \cellcolor{orange!8} \textbf{55.2}
& \cellcolor{orange!8} \textbf{56.0}
& \cellcolor{orange!8} \textbf{53.1}
& \cellcolor{orange!8} 47.7
& \cellcolor{orange!8} \textbf{54.0} \\
\bottomrule
\end{tabular}%
}
\label{tab:smplx_smplh}
\end{table}
\subsection{Use of a different encoder: OmniFit}
\label{sec:omnifit}
To further assess the generality of our framework beyond the ArtEq encoder, we evaluate it on top of OmniFit~\citep{cai2026omnifit}, a recent multi-modal, scale-agnostic method that fits the SMPL-X body model to 3D human assets. We treat its reconstruction as the source mesh $\mathcal{S}=\mathrm{Dec}^{\mathcal{L}}(C)$ and apply our refinement methods to recover the residual geometry, without any change to their architecture.

The learning-based methods (Latent and Vertex-VaRefine) are trained on the OmniFit outputs of the BML\_movi,
while ElaOpt, being optimization-based, requires no training; all methods are
then evaluated on the remaining subsets (GRAB, TCD, HumanEva, EKUT, SFU), so that the reported scores measure generalization to
unseen sequences. Results are given in Table~\ref{tab:omnifit_refinement},
using the vertex-to-vertex (V2V) and the Chamfer distance (CD), both in
centimeters. Interestingly, we obtain even better improvements than for ArtEq, showcasing the generalizability of our method to both a different latent space (SMPL-X) and a completely different decoder.

\begin{table}[t]
\centering
\caption{Per-dataset evaluation on the AMASS subsets. OmniFit gives the
initial reconstruction; we report the error after each refinement method and
the relative gain over OmniFit. Lower is better; all values in cm.}
\label{tab:omnifit_refinement}
\resizebox{\columnwidth}{!}{%
\begin{tabular}{l|l|c|ccccc}
\toprule

\textbf{Method} 
& \textbf{Metric} 
& \textbf{AMASS} 
& \textbf{GRAB} 
& \textbf{TCD} 
& \textbf{HumanEva} 
& \textbf{EKUT} 
& \textbf{SFU} \\

\midrule

\multirow[c]{2}{*}{OmniFit}

& V2V
& \cellcolor{gray!8} 
1.181 & \cellcolor{gray!8} 1.130
& \cellcolor{gray!8} 1.112
& \cellcolor{gray!8} 1.167
& \cellcolor{gray!8} 1.592
& \cellcolor{gray!8} 1.162 \\

& CD
& \cellcolor{gray!8} 0.649
& \cellcolor{gray!8} 0.641
& \cellcolor{gray!8} 0.706
& \cellcolor{gray!8} 0.616
& \cellcolor{gray!8} 0.727
& \cellcolor{gray!8} 0.655 \\

\midrule
\multirow[c]{4}{*}{Latent-VaRefine}

& V2V
& \cellcolor{blue!8} 
0.822 & \cellcolor{blue!8} 
0.825 & \cellcolor{blue!8} 
1.092 & \cellcolor{blue!8} 
0.853 & \cellcolor{blue!8} 
0.822 & \cellcolor{blue!8} 0.868 \\

& Gain (\%)
& \cellcolor{blue!8} 
30.4
& \cellcolor{blue!8} 
27.0
& \cellcolor{blue!8} 
1.8 & \cellcolor{blue!8} 
26.9
& \cellcolor{blue!8} 
48.4
& \cellcolor{blue!8} 
25.3\\

\cmidrule(lr){2-8}

& CD
& \cellcolor{orange!8}
0.533
& \cellcolor{orange!8}
0.529
& \cellcolor{orange!8}
0.700
& \cellcolor{orange!8} 0.482
& \cellcolor{orange!8}
0.520
& \cellcolor{orange!8}0.572\\

& Gain (\%)
& \cellcolor{orange!8} 17.9
& \cellcolor{orange!8} 17.5 
& \cellcolor{orange!8} 0.8
& \cellcolor{orange!8} 21.8
& \cellcolor{orange!8} 28.5
& \cellcolor{orange!8}12.7  \\
\midrule

\multirow[c]{4}{*}{Vertex-VaRefine}

& V2V
& \cellcolor{blue!8} \textbf{0.342}
& \cellcolor{blue!8} \textbf{0.328}
& \cellcolor{blue!8} \textbf{0.400}
& \cellcolor{blue!8} \textbf{0.311}
& \cellcolor{blue!8} \textbf{0.339}
& \cellcolor{blue!8} \textbf{0.419}\\

& Gain (\%)
& \cellcolor{blue!8} \textbf{71.0}
& \cellcolor{blue!8} \textbf{71.0}
& \cellcolor{blue!8} \textbf{64.0}
& \cellcolor{blue!8} \textbf{73.4}
& \cellcolor{blue!8} \textbf{78.7}
& \cellcolor{blue!8} \textbf{63.9} \\

\cmidrule(lr){2-8}

& CD
& \cellcolor{orange!8} \textbf{0.274}

& \cellcolor{orange!8} 0.253
& \cellcolor{orange!8} 0.329
& \cellcolor{orange!8} \textbf{0.255}
& \cellcolor{orange!8} \textbf{0.274}
& \cellcolor{orange!8} 0.323  \\

& Gain (\%)
& \cellcolor{orange!8} \textbf{57.8}
& \cellcolor{orange!8} 60.5
& \cellcolor{orange!8} 53.4
& \cellcolor{orange!8} \textbf{58.6}
& \cellcolor{orange!8} \textbf{62.3}
& \cellcolor{orange!8} 50.7 \\

\midrule

\multirow[c]{4}{*}{ElaOpt}

& V2V
& \cellcolor{blue!8} 0.967
& \cellcolor{blue!8} 0.954
& \cellcolor{blue!8} 0.902
& \cellcolor{blue!8} 0.934
& \cellcolor{blue!8} 0.982
& \cellcolor{blue!8} 0.930\\

& Gain (\%)
& \cellcolor{blue!8} 
18.1 & \cellcolor{blue!8} 15.6
& \cellcolor{blue!8} 18.9
& \cellcolor{blue!8} 20.0
& \cellcolor{blue!8} 38.3
& \cellcolor{blue!8} 20.0 \\

\cmidrule(lr){2-8}

& CD
& \cellcolor{orange!8} 0.289
& \cellcolor{orange!8} \textbf{0.245}
& \cellcolor{orange!8} \textbf{0.308}
& \cellcolor{orange!8} 0.281
& \cellcolor{orange!8} 0.381
& \cellcolor{orange!8} \textbf{0.311} \\

& Gain (\%)
& \cellcolor{orange!8} 55.5
& \cellcolor{orange!8} \textbf{61.8}
& \cellcolor{orange!8} \textbf{56.4}
& \cellcolor{orange!8} 54.4
& \cellcolor{orange!8} 47.6
& \cellcolor{orange!8} \textbf{52.5} \\

\bottomrule
\end{tabular}%
}
\end{table}

%\textcolor{red}{I think this next paragraph could use 2-3 more to better explain what is tested.}

\subsection{Robustness to Target Mesh Variations}
\paragraph{Mesh Resolution} To test that all three refining methods are robust to remeshing, we apply them to both upsampled and downsampled versions of 100 random
target meshes from the SFU dataset. We report the results in Table~\ref{tab:resolution}, which confirms our statement.

\begin{table}[]
\centering
\small
\setlength{\tabcolsep}{5pt}

\caption{Robustness to mesh resolution. Relative improvement (\%) in asymmetric
Chamfer Distance (CD) over ArtEq on 100 target meshes from the SFU dataset,
for the original, downsampled and upsampled targets. The methods are applied
as trained on SMPL targets.}
\begin{tabular}{l|ccc}
\toprule
\textbf{Method} & \textbf{Original} & \textbf{Downsampled} & \textbf{Upsampled} \\
\midrule
Latent-VaRefine & 17.8 & 18.5 & 17.5 \\
Vertex-VaRefine & 54.7 & 46.3 & 46.9 \\
ElaOpt          & 65.7 & 66.1 & 65.8 \\
\bottomrule
\end{tabular}
\label{tab:resolution}
\end{table}

\begin{figure}
    \centering
    \includegraphics[width=.9\linewidth]{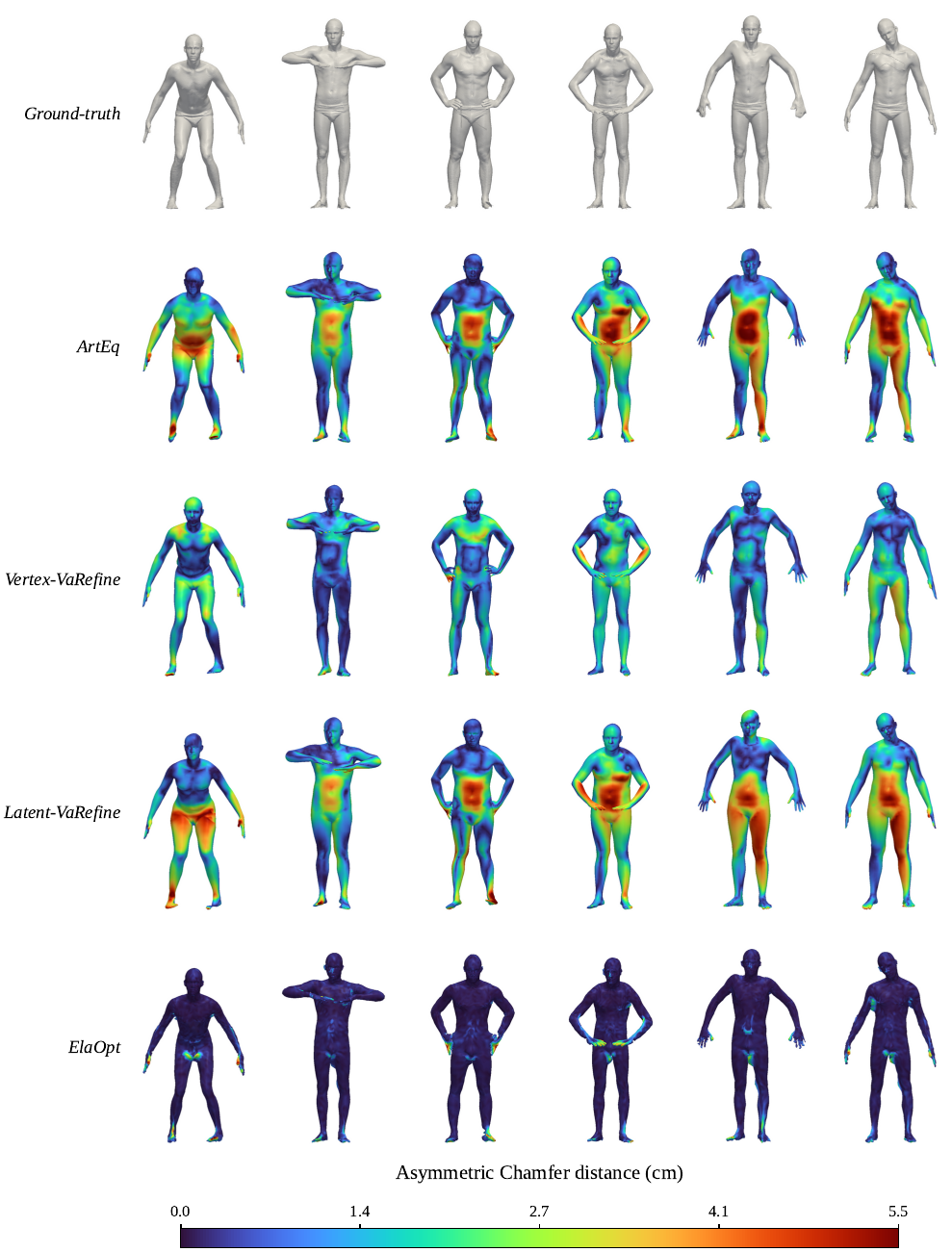}
    \caption{Qualitative comparison on 10 raw FAUST scans. Color maps show the
    per-vertex asymmetric Chamfer distance to the ground-truth scan surface, in
    centimeters (cm).}
    \label{fig:faust_heatmap}
\end{figure}

\paragraph{Raw Scan Targets} We test our methods on 10 raw
scans from FAUST, i.e.,\ targets completely outside of any parametric body model. The scans were chosen
so that ArtEq gives a relatively accurate pose, before checking the improvements
of the proposed methods (Table~\ref{tab:faust_scans}, Figure~\ref{fig:faust_heatmap}). In this case, only (asymmetric) Chamfer distance makes sense as a measurement. As expected, Latent-VaRefine cannot improve the result by much, as it is constraiend to stay within the SMPL space. On the other hand, both vertex-based methods give marked improvement, with the Optimization-based method logically being the most generalizable to this far-from-training situation.

\begin{table}[H]
\centering
\small
\setlength{\tabcolsep}{5pt}
\caption{Generalization to non-SMPL targets. Relative improvement (\%) in
asymmetric Chamfer Distance (CD) over the ArtEq prediction on 10 raw scans from
FAUST. The scans were chosen so that ArtEq gives a relatively accurate pose,
before evaluating any of the proposed methods.}
\resizebox{\columnwidth}{!}{%
\begin{tabular}{l|ccc}
\toprule
\textbf{Metric} & \textbf{Latent-VaRefine} & \textbf{Vertex-VaRefine} & \textbf{ElaOpt} \\
\midrule
CD Gain (\%) & 3.9 & 29.1 & \textbf{65.2} \\
\bottomrule
\end{tabular}%
}
\label{tab:faust_scans}
\end{table}

\section{Limitations and Future work}
%\textcolor{red}{This section does not describe future works or how this could be improved}

\paragraph{Individual Weakness of Each Approach} %A first limitation is the individual weaknesses of the three methods: 
From the results of the previous sections, one can conclude that Latent-VaRefine preserves the latent representation but is less accurate, though it can be easily combined with the two other frameworks to effect; Vertex-VaRefine achieves the lowest vertex-to-vertex errors; while ElaOpt yields the best Chamfer distances and is extendable to clothed meshes (see Appendix \ref{sec:clothes}), at the cost of a substantially slower inference and potentially more irregular output meshes. Combining a learning based initialization with few subsequent optimization steps is a natural direction for future work. Another common limitation to all proposed methods is the assumption that the source mesh produced by the base latent code retrieval method is not too far from the ground truth, especially in terms of the predicted pose. This could possibly be improved by adapting Latent-VaRefine to capture bigger changes, which we intend to investigate.     

% \paragraph{Encoder Limitation} 
% One of the main limitations of all methods is that they require the source mesh generated by the chosen latent code retrieval method to be suitably close to the target. If the pose retrieved is too far from the ground truth, the results of the vertex-based methods will numerically reduce the error, but look clearly bad (e.g., collapsing the hands). Indeed, we tried a few other encoders such as IPnet~\citep{IPNetBhatnagar20a}, but the results were mediocre because of this. This could possibly be improved by adapting Latent-VaRefine to capture bigger changes, and combining it to the other methods, which we intend to investigate. 

\paragraph{Shape and Motion Generation} In their current form, all three methods allow for animation of the refined shape using the pose parameter of the latent space from which the source shape is obtained. This could be used in conjunction with motion generation algorithms such as ARDY~\citep{ARDY} to animate general body shapes. On the other hand, the generation of more general human body shapes using generative methods could be possible, though it will probably require adapting the Vertex-VaRefine approach to allow for a useful latent space on which to work. This is another topic of interest as a natural follow-up to this work.

\paragraph{Clothed Human Shapes and Scans} While the addition of clothes is an obvious next step for our vertex-based frameworks, preliminary results have only showed good predictions for ElaOpt so far; we provide some of those results and examples in Appendix \ref{sec:clothes}. We intend to pursue this both with the optimization framework and by a possible change of paradigm in Vertex-VaRefine.

%there is no latent space for the Vertex-based methods on which to train any kind of the usual shape generation methods. In the case of Vertes-VaRefine, one could possibly build

\section{Conclusion}
This work proposed different approaches for the post-correcting and
refining of latent space representations for human body scans estimated
by currently available latent code retrieval algorithms. We specifically
address the problem of improving the code prediction via a correction
within the considered latent space, which we complement with two
different methods that operate at the vertex level and allow the capture
of missing details beyond the limitations of the latent model itself.

The results reveal a practical trade-off between preserving the
original parametric representation and improving geometric fidelity.
Latent-VaRefine retains the model parameterization, Vertex-VaRefine
provides strong vertex correspondence accuracy, and ElaOpt favors
surface fidelity through direct geometric optimization. Their
complementarity supports choosing the refinement strategy according
to the requirements of the downstream application.
%A remaining limitation is the individual weaknesses of the three methods: Latent-VaRefine preserves the latent representation but is less accurate, though it can be easily combined with the two other frameworks to effect; Vertex-VaRefine achieves the lowest vertex-to-vertex errors; while ElaOpt yields the best Chamfer distances and is extendable to clothed meshes, at the cost of a substantially slower inference and a messier mesh. Combining a learning based initialization with some short following optimization steps, combining our frameworks, is a natural direction for future work.

\subsection{Acknowledgments}

Chaima Zouaoui, Sylvain Arguillère, and Mohamed Daoudi were supported by the project 4DSHAPE ANR-24-CE23-5907 of the French National Research Agency (ANR). M. Bauer and J. Cerqueira were supported by the National Science Foundation through grant 2426549. N. Charon and E. Hartman were supported by the National Science Foundation through grant 2426550. 

\subsection{Statement on AI usage}

During the preparation of this manuscript, the authors used AI tools solely for language editing to improve clarity and readability. AI tools were not used to generate scientific content, conduct analyses, or draw conclusions. The authors reviewed and verified all edits and take full responsibility for the content of the published article.

\printcredits

\bibliographystyle{cas-model2-names}

\bibliography{cas-refs, amass}
\newpage

\appendix
%\chapter{Supplementary Material}
\noindent
\textbf{\Large{Appendix}}

\section{Distances Between Surface Meshes}

\label{sec:distances}
In this section, we go over the different notions of distances between surface meshes (registered or not) that are used throughout the paper, both to quantify reconstruction errors and as loss functions for the proposed neural network models. In all the following, we consider two meshes with vertex sets
$\mathbf{V}=\{\mathbf{x}_i\}_{i=1}^{M}$ and
$\mathbf{V}'=\{\mathbf{x}'_j\}_{j=1}^{N}$.

\paragraph{Vertex-to-Vertex Distance (V2V)}
When the two meshes share the same topology and their vertices are in one-to-one correspondence, then the vertex-to-vertex error measures the average Euclidean distance between corresponding vertices:
\[\mathmakebox[\linewidth][c]{
\mathrm{V2V}(\mathbf{V},\mathbf{V}') =
\frac{1}{M} \sum_{i=1}^{M} \left\|\mathbf{x}_i - \mathbf{x}'_i\right\|_2 
}\]
where $\|\cdot\|_2$ denotes the Euclidean norm. Throughout the article, results using this measurement will be reported in centimeters.

%\paragraph{Symmetric Chamfer Distance (SCD)}
%The symmetric Chamfer Distance measures the geometric discrepancy between two point sets by averaging nearest-neighbor distances in both directions:
%\[\mathmakebox[\linewidth][c]{
%\begin{aligned}
%\mathrm{SCD}(\mathbf{V},\mathbf{V}')=\;&
%\frac{1}{M}\sum_{i=1}^{M}\min_{1\le j\le N}\left\|\mathbf{x}_i-\mathbf{x}'_j\right\|_2\\
%+\;&
%\frac{1}{N}\sum_{j=1}^{N}\min_{1\le i\le M}\left\|\mathbf{x}'_j-\mathbf{x}_i\right\|_2
%\end{aligned}
%}\]
%Unlike the V2V metric, the Chamfer Distance does not require any vertex correspondences or shared topology, which makes it well suited to comparing unregistered surfaces.
% This symmetric distance is used to decide whether to keep the refined reconstruction or revert to the initial reconstruction. The refinement is accepted only if the symmetric Chamfer Distance is reduced. ElaOpt also relies on this criterion during the optimization process.

\paragraph{Asymmetric Chamfer Distance (CD)} 
A common alternative to the above is the \emph{asymmetric} Chamfer Distance defined by
\[\mathmakebox[\linewidth][c]{
\mathrm{CD}(\mathbf{V},\mathbf{V}')
=\frac{1}{N}\sum_{j=1}^{N}\min_{1\le i\le M}\left\|\mathbf{x}'_j-\mathbf{x}_i\right\|_2}\]
The asymmetric Chamfer Distance measures how well the surface $\mathbf{V}$ covers $\mathbf{V}'$, and does not require vertex correspondences or shared topology. Being one-directional, it is robust to case where the target mesh is incomplete or contains holes. One can also define the \emph{symmetric} Chamfer distance by simply summing the two asymmetric distances i.e. \(\mathrm{SCD}(V,V')=\mathrm{CD}(V,V')+\mathrm{CD}(V',V)\). Results for either version of the Chamfer distance will be reported in centimeters.

\paragraph{Vertex-to-Vertex Supremum}
Under the same correspondence assumption as the V2V error (Section~\ref{sec:distances}), replacing the average with a maximum gives the worst-case point-wise deviation:
\[\mathmakebox[\linewidth][c]{
\mathrm{V2V}_{\sup}(\mathbf{V},\mathbf{V}') =
\max_{1\le i\le M} \left\|\mathbf{x}_i - \mathbf{x}'_i\right\|_2 
}\]
Unlike the mean V2V error, this metric is highly sensitive to localized reconstruction failures, such as a single poorly reconstructed limb or extremity. Results are reported in centimeters.

\paragraph{Hausdorff Distance}
The Hausdorff distance is a symmetric, worst-case counterpart to the Chamfer distance (Section~\ref{sec:distances}): rather than averaging nearest-neighbor distances, it retains their maximum in each direction, then takes the larger of the two:
\[\mathmakebox[\linewidth][c]{
\begin{aligned}
d_{H}(\mathbf{V},\mathbf{V}')=\;&
\max\Big(\max_{1\le i\le M}\min_{1\le j\le N}\left\|\mathbf{x}_i-\mathbf{x}'_j\right\|_2,\\
&\max_{1\le j\le N}\min_{1\le i\le M}\left\|\mathbf{x}'_j-\mathbf{x}_i\right\|_2\Big)
\end{aligned}}
\]
Like the Chamfer distance, it does not require vertex correspondence, but it is considerably more sensitive to outliers and localized failures. Results are reported in centimeters.

\paragraph{Varifold Distances}
The varifold distance $d^{\mathrm{Var}}_\sigma$ is defined in Eq.~\eqref{eq:var_dist} (Section~\ref{sec:varigrad}) and compares both the positions and the orientations of the surfaces being compared. We consider its two standard variants: the \emph{oriented} distance, obtained with the orientation kernel $\Gamma(N\cdot N') = e^{(N\cdot N'-1)/\tau^2}$ used throughout the paper, and the \emph{unoriented} distance, obtained with $\Gamma(N\cdot N') = (N\cdot N')^2$. The latter depends only on the square of the inner product between normals and is therefore invariant to a flip of the normal orientation of either mesh.

\section{SMPL to SMPL-H and SMPL-X Converters}
\label{app:converter}
\begin{figure}[]
    \centering
    \includegraphics[width=.7\linewidth]{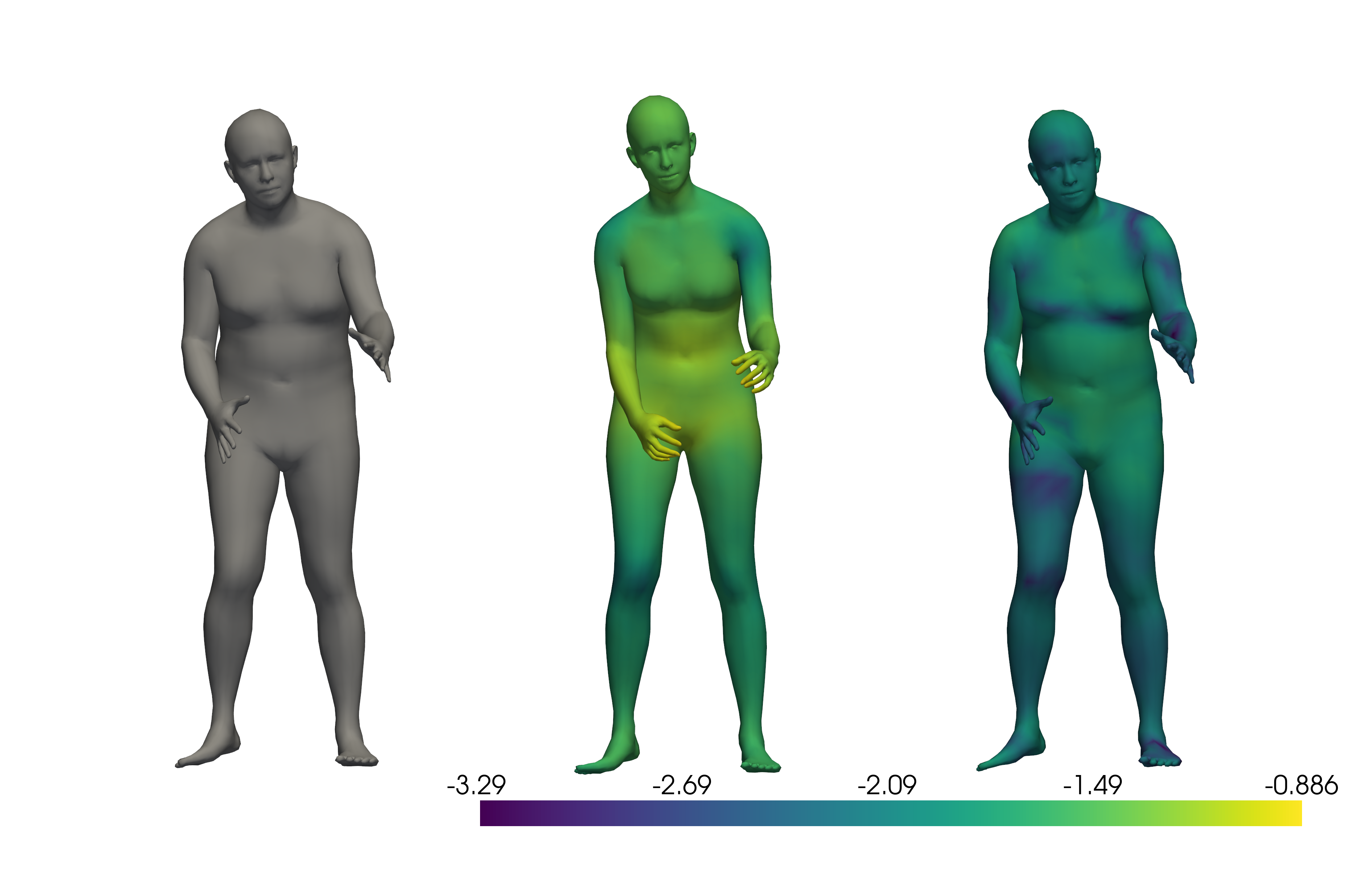}
    \caption{The above is an SMPL mesh (left) in gray along with the results of naive parameter reuse into SMPL-X (center) and an SMPL-X mesh (right) that is the result of an optimization on the SMPL-X parameters to minimize MSE with the ground truth. As SMPL and SMPL-X have different topologies, this MVE and V2V distance are calculated after first applying SMPL's supplied matrix to the mesh to convert its mesh topology to that of SMPL. The plotted error is V2V distance in meters under a log of base 10.}
    \label{fig:optimizedcoef}
\end{figure}

The parameters of SMPL latent codes can be used naively as SMPL-X or SMPL-H parameters. One simply matches up shape (\(\shapecoeff\)) parameters and assigns the pose (\(\posecoeff\)) parameters of one latent space to the relevant pose parameters of the other. The immediate issue is the simple architectural differences between the different latent spaces. SMPL offers 6 coefficients for hands, but SMPL-X and SMPL-H offer as many as 90, 45 for either hand, if one does not use PCA. SMPL-X adds expression coefficients for the face and parameters for jaw and eye position. Even if one chooses, as a matter of convention, to simply discard hand and expression information and match only the common joints and the shape parameters, the result is not the identity. That is, the exact mesh encoded by a set of shape (\(\shapecoeff\)) and joint (\(\posecoeff\)) parameters depends on the decoder used even if one restricts to parameters held in common. See Figure \ref{fig:optimizedcoef} for a typical example.

To avoid retraining ArtEq for SMPL-X or performing long intensive computations to find optimal matchings between the thousands of relevant SMPL and SMPL-X parameters, we resort to training a simple neural network for the task. The fact that makes this workable is that while the meshes encoded by SMPL and SMPL-X by a common set of parameters will differ, they do not differ extremely. The poses are nearly right. Additionally, there is a matrix, which we will here denote \(\mathcal{M}\), provided by the SMPL(-X) ecosystem to convert a generated SMPL-X mesh into one with the topology of an SMPL mesh. This allows us to train an autoencoder to find parameters correspondences with MSE loss:
\[\operatorname{Loss}(C')= \operatorname{MSE}(\mathrm{Dec}^{\mathrm{SMPL}}(C), \mathcal{M}\mathrm{Dec}^{\mathrm{SMPLX}}(C')).\]
These losses are computed after an area-weighted centroid based centering, as SMPL and SMPL-X do not have a common center.

\begin{figure}
    \centering
    
    \fbox{\includegraphics[width=0.97\linewidth]{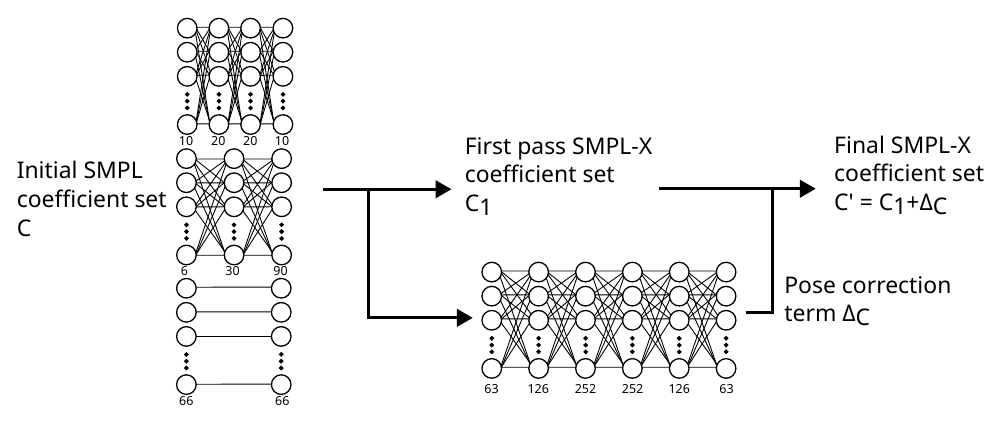}}
    \caption{Overview of the autoencoder's architecture. The upper left subnetwork is for $\shapecoeff$ conversion, the center left subnetwork is for hands. Together their outputs form an initial guess, with the pose and orientation retained naively. The pose parameters are then corrected by the larger network in the center of the figure.}
    \label{fig:converterarch}
  \end{figure}

\begin{figure}[]
    \centering
     
    \includegraphics[width=.7\linewidth]{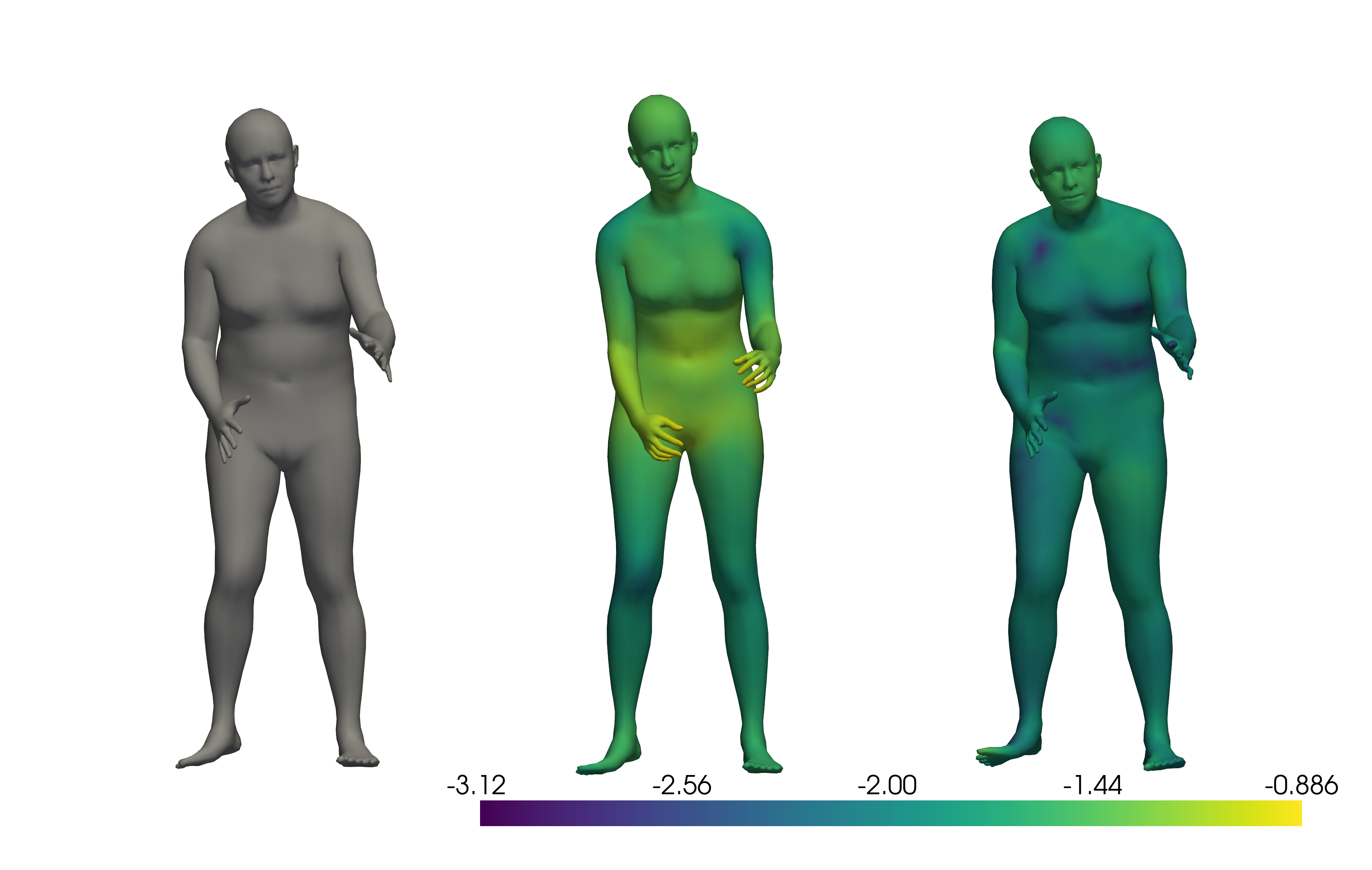}
   \caption{The above is an SMPL mesh (left) in gray along with the results of naive parameter reuse in SMPL-X (center) and the SMPL-X mesh (right) that produced by the autoencoder. The result here is nearly as good as the optimized result above in Figure \ref{fig:optimizedcoef}. The plotted error is V2V distance in meters under a log of base 10.}
    \label{fig:convertedXcoef}
\end{figure}

Naively attempting to train with the above loss with a single large network  appears to be slow and prone to local optima. More effective is to train in three stages. We trained the model to convert SMPL latent codes with 10 shape parameters into SMPL-X latent codes with 10 shape parameters. The SMPL latent code is 82 parameters: 3 parameters corresponding to a global orientation, 63 corresponding to non-hand joints, 6 for SMPL's simple hand joints, and 10 shape parameters. The SMPL-X latent code is 166 parameters: 3 parameters corresponding to a global orientation, 63 corresponding to non-hand joints, 90 for SMPL-H style hands, and 10 shape parameters. The first stage is a pair of MLPs, one MLP converting the 6 SMPL hand joint parameters into 90 SMPL-H style parameters and one MLP converting the 10 shape parameter from SMPL parameters to SMPL-X parameters. Training this stage alone will find parameters such that the shape will better match than naive parameter matching. The body pose can then be corrected by training an MLP to find a correcting term for the pose parameters. Finally, once both are roughly correct, the end to end combination of the two stages can be trained together for a final performance increase. The overall architecture is shown in Figure~\ref{fig:converterarch} and an example result can be seen in Figure~\ref{fig:convertedXcoef}

\begin{figure}[]
    \centering
    
    \includegraphics[width=.7\linewidth]{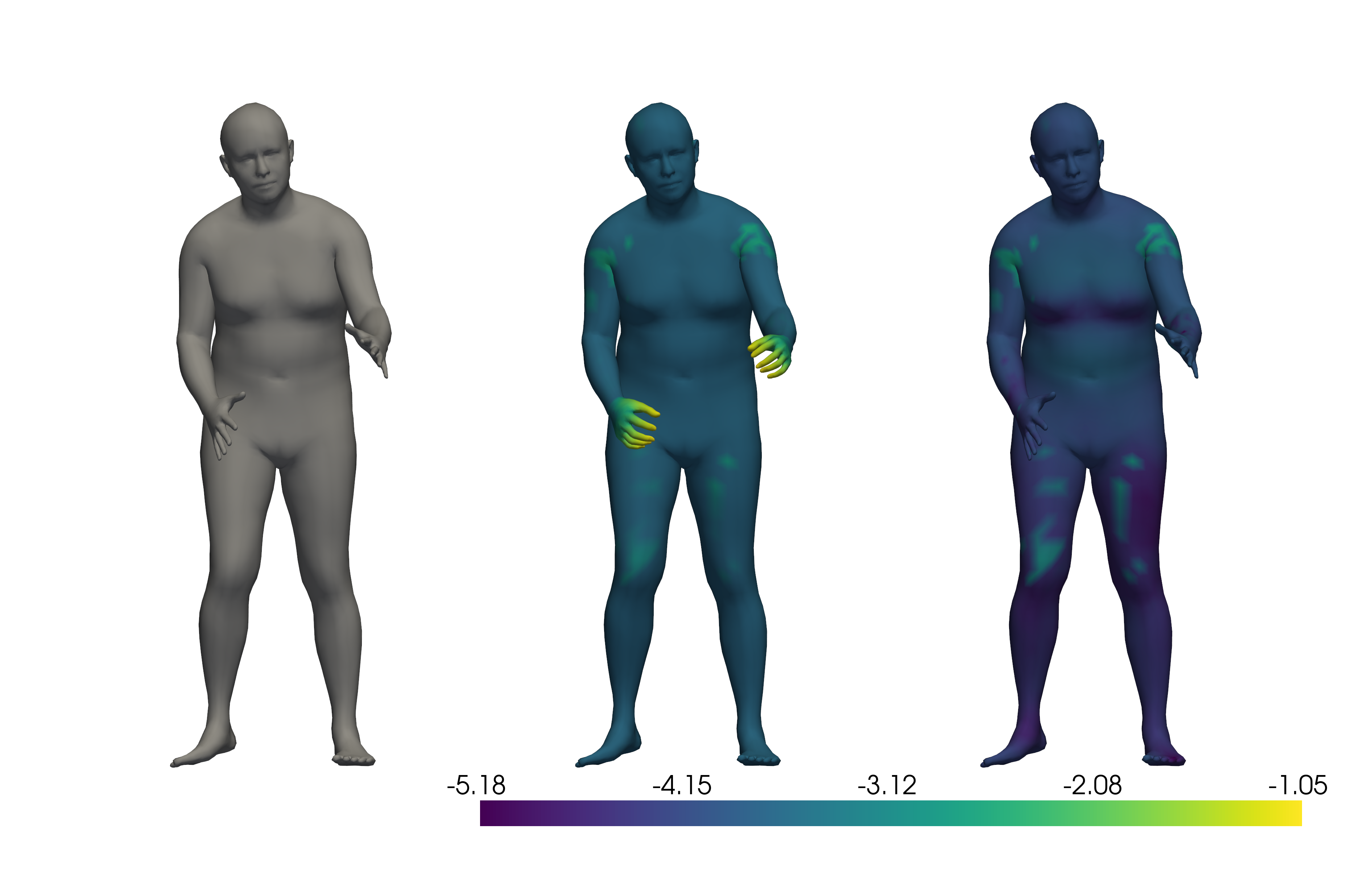}
    \caption{Similar architecture may be trained to allow for conversion to SMPL-H. Note here the naive result is much closer to correct and the converter largely serves to fix the incorrect hands.}
    \label{fig:convertedHcoef}
\end{figure}

We can perform a nearly identical process to create a converter from SMPL parameters to SMPL-H parameters. As we did not use expression or jaw parameters above, we have the same 166 parameters for a 10-shape parameter SMPL-H latent code. Unlike in the SMPL-X case, there is no difference of mesh topology, so our loss is simply
\[\operatorname{Loss}(C')= \operatorname{MSE}(\mathrm{Dec}^{\mathrm{SMPL}}(C), \mathrm{Dec}^{\mathrm{SMPLH}}(C')).\]
 Naive parameter reuse is closer to the identity for SMPL-H. It largely loses only hand information. Due to this, the converter performs even better than in the SMPL-X case as can be seen in Figure~\ref{fig:convertedHcoef}.

\section{Alternative Backbones for Vertex-VaRefine}
\label{app:backbones}
The choice of DiffusionNet as the backbone of Vertex-VaRefine results from the ablation study of Section~\ref{sec:ablation_backbones}. We describe here the two alternative backbones evaluated in place of DiffusionNet, as well as the two decoding heads used to obtain $D$: a direct per-vertex displacement and a Neural Jacobian Field.

\paragraph{PoissonNet}
PoissonNet~\citep{maesumi2025poissonnet} is an intrinsic mesh neural network that replaces the spectral diffusion operator of DiffusionNet~\citep{sharp2022diffusionnet} with a differentiable Poisson equation solving as its mechanism for information propagation across the surface. Given per-vertex scalar features $x \in \mathbb{R}^{V \times C}$, each Poisson block proceeds in three steps: (i) the intrinsic gradient operator $G \in \mathbb{R}^{2F \times V}$ maps vertex features to per-face tangent-plane gradients, $\nabla x = Gx$, represented as complex numbers in a local 2D basis per face; (ii) a complex-valued MLP transforms these tangent-vector features while preserving the $SO(2)$-equivariance of the local frame; (iii) a Poisson equation $Lu = \operatorname{div}(A \cdot \text{grads})$ is solved in closed form via a pre-factorized (Cholesky) cotangent Laplacian $L$~\citep{nicolet2021large} -- one factorization per mesh, reused across all channels and both forward/backward passes -- to obtain a globally smoothed vertex signal $u$. The block output is $\mathrm{MLP}([x, u]) + x$. Stacking $N$ such blocks after a linear input projection gives the PoissonNet backbone; a final linear head maps $C_{\text{width}}$-dimensional vertex features to the task output.

\paragraph{SpiralNet}
SpiralNet++~\citep{gong2019spiralnet++} offers an attractive compromise: it
defines a fast, anisotropic mesh convolution by following spiral sequences of neighbouring vertices on a mesh of fixed topology. In the case where the source shape has a fixed mesh template, like in ours (the source mesh $\mathcal{S}$ is always decoded from a given latent space and therefore has consistent mesh structure), the spirals can be precomputed once and reused across the dataset, leading to an efficient and topology-aware encoder particularly suited to body meshes. The final output then needs to be as detailed as possible, so we forego the usual downsampling of SpiralNet. This lets us directly obtain a per-vertex processed feature $z=(z_1,\dots,z_V)\in \mathbb{R}^{K\times V}$. Finally, to prime this code for the next module, we turn it into \textit{per-face} feature $Z=(Z_1,\dots,Z_F)\in \mathbb{R}^{K\times F}$. For each face index $f=1\dots F$ of the mesh, $Z_f$ is the average of the three codes $z_is$ attached to the three vertices of that face.

\paragraph{Decoding Head: Direct Displacement vs.\ Neural Jacobian Fields}
For DiffusionNet and PoissonNet, the backbone directly outputs the per-vertex T-pose displacement $D=\mathrm{Net}(\phi)\in\mathbb{R}^{V\times 3}$ through a final linear layer, with no further post-processing.
For SpiralNet++, directly regressing $D$ from per-vertex features produces noisy and spatially inconsistent corrections. To enforce smoothness without sacrificing expressivity, we instead adopt a slightly modified version of the Neural Jacobian Field~\citep{NeuralJacobianFieldACMTG2022} formulation on top of the SpiralNet++ encoder. Rather than predicting the displacement directly, the network outputs a per-triangle $3{\times}3$ Jacobian matrix $J_f=\operatorname{MLP}(Z_f,c_f)$, $f=1,\dots,F$, from the per-face encoder features $Z_f$ and face centers $c_f$ introduced above. The columns of $J_f$ are interpreted as the surface gradient of each coordinate of the T-pose correction. The deformation is then recovered by solving the Poisson system
\[\mathmakebox[\linewidth][c]{
   L_{\mathcal{S}_0}\,\Delta = \mathcal{A}_{\mathcal{S}_0}\,
   \nabla^{T}_{\mathcal{S}_0}\, J_f ,
}\]
where $L_{\mathcal{S}_0}$ is the surface Laplacian, $\mathcal{A}_{\mathcal{S}_0}$ the diagonal matrix of face areas, and $\nabla^{T}$ the transpose of the surface gradient operator, all computed once on a fixed reference template $\mathcal{S}_0$ (the template SMPL mesh) shared by every training sample. Since $\mathcal{S}$, and hence $\bar{\mathcal{S}}$, always share the SMPL topology, this Poisson system is factorized only once, independently of the training set size, which keeps memory and training time tractable. Unlike the original, non-equivariant formulation, the resulting correction $\Delta$ is here added to the per-sample T-pose shape rather than to $\mathcal{S}_0$ itself, $D \doteq \Delta$ with $T=\bar{\mathcal{S}}+\Delta$ -- this is what makes this decoding head compatible with the pose-equivariant pipeline described above. This approach has two advantages: it introduces a regularization prior on the deformation appropriate for body geometry, and it preserves the intrinsic differential structure of the source surface, so fine-scale features are not distorted during refinement.

\section{Clothed Human Scans and Scan Animation}\label{sec:clothes}

In a manner similar to \emph{SMPL+D} and \cite{IPNetBhatnagar20a}, per vertex correction preserves the original joint rigging and skinning weights allowing the reconstruction be reposed or animated. Additionally, unlike methods constrained to the latent space's representative limits, these vertex corrections can be applied to capture clothing or other details outside the latent space constraints. These are items we will explore in future work, and we present in this section some promising early proof-of-concept efforts.

\begin{figure}
    \centering    
    
    \includegraphics[width=1\linewidth]{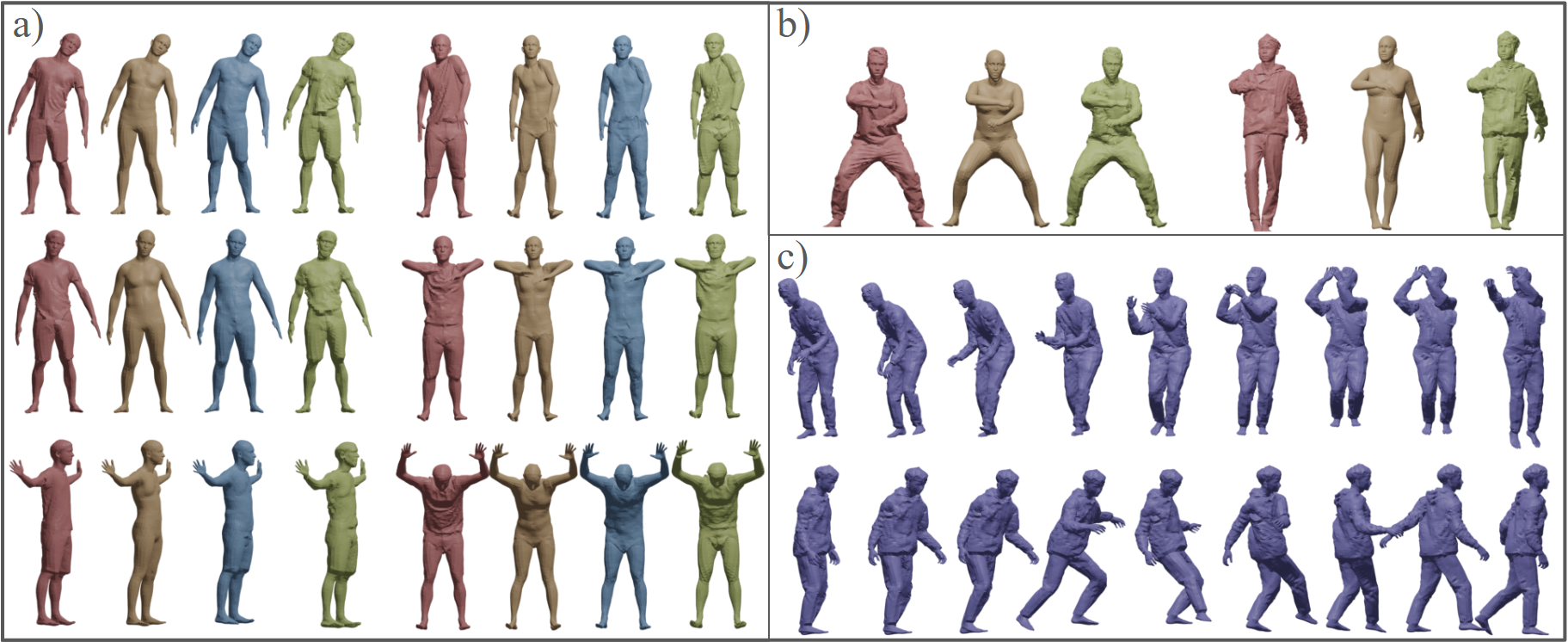}
    \caption{ In a), we present six examples from the CAPE Dataset. The ground truth clothed meshes are in red, the SMPL approximations are presented in yellow, the refined meshes via Vertex-Varefine are presented in blue, and the refined meshes produced by ElaOpt are presented in green.  In  b), we present two examples from the THumans dataset. Again, the raw scans are displayed in red, the SMPL approximations are displayed in yellow, and the reconstruction using ElaOpt is displayed in green. In c), we display the transfer of a motions from the CMU dataset to the clothed meshes using the outputs of the ElaOpt method from b).}
    \label{fig:ClothesScanResults}
\end{figure}

Figure~\ref{fig:ClothesScanResults} a-b), presents qualitative reconstructions of scans from the CAPE~\citep{CAPE:CVPR:20} and THuman~\citep{tao2021function4d} datasets. Starting from minimally clothed SMPL meshes, each method predicts deformation fields that recover the geometry of the target clothed scans. Quantitatively, the Vertex-VaRefine framework reduces the vertex-to-vertex Error (V2V) from \(1.74\,\mathrm{cm}\) to \(1.61\,\mathrm{cm}\), corresponding to a relative improvement of approximately \(7.5\%\). Meanwhile, the ElaOpt method reduces the V2V to \(1.41\,\mathrm{cm}\) corresponding to a relative improvement of \(19\%\). Qualitatively, VaRefine captures the overall clothing geometry, but not enough to convincingly dress the source mesh. Fine-scale cloth wrinkles and high-frequency surface details remain challenging. ElaOpt on the other hand, yields much better-looking results, both in terms of large scale geometry and wrinkles.
%~\citep{CAPE:CVPR:20} with both the Vertex-VaRefine model and ElaOpt, either of which allow leaving of the SMPL latent space. CAPE provides dynamic 3D clothed human meshes together with minimally clothed SMPL body meshes in point-to-point vertex correspondence, making it well suited for deformation-based mesh refinement. The dataset contains realistic garment deformations from clothing such as shirts, jackets, skirts, and shorts, resulting in substantially more challenging reconstruction scenarios than minimally clothed body datasets. We adopt a subject-independent% and clothing-independent evaluation protocol, reserving unseen subjects and the \textit{shortshort} clothing category exclusively for testing. In total, 8,674 meshes are used for training. Additionally, we qualitatively evaluate the ElaOpt method on clothed meshes from the the THumans dataset~\citep{tao2021function4d}
%For more challenging examples, we evaluate the ElaOpt method on clothed scans from the THumans dataset.

Since the ElaOpt method deforms the shape in the normal direction of the SMPL representation, these deformations can be applied to a sequence of pose parameters to produce 4D sequences of clothed meshes. A qualitative evaluation of the ElaOpt method's ability to capture clothing details and an example of a resulting transfer of clothes to a 4D sequences are presented in Figure~\ref{fig:ClothesScanResults} c).%, and as animations in the \textbf{supplementary material}.

\section{Additional Results}
\label{app:Additional}

This appendix provides the complete quantitative evaluation of Latent-VaRefine, Vertex-VaRefine, ElaOpt, Latent-VaRefine + ElaOpt, and Latent-VaRefine + Vertex-VaRefine on all evaluated AMASS sub-datasets. Tables \ref{tab:amass_subdatasets_supp_MVE} and \ref{tab:amass_subdatasets_supp_ACham} list the detailed reconstruction errors for the remaining AMASS sub-datasets not reported in Section \ref{ssec:results_comparison} of the main paper. 

\begin{table*}[]
\centering
\setlength{\fboxsep}{3pt}
\setlength{\fboxrule}{0.6pt}
\makebox[\textwidth][c]{%
\fbox{%
\begin{minipage}{\textwidth}
\centering
\small
\setlength{\tabcolsep}{2pt}
\renewcommand{\arraystretch}{0.9}
\caption{Detailed evaluation on additional AMASS sub-datasets (SMPL target). We report \textbf{vertex-to-vertex} (V2V, in cm) for ArtEq, Latent-VaRefine, ElaOpt, and Vertex-VaRefine together with the corresponding relative improvements.}
\label{tab:amass_subdatasets_supp_MVE}
\begin{tabular}{l|cccccc}
\toprule
Dataset & ACCAD & BML\_rub & CMU & CNRS & DFAUST & DanceDB \\
\midrule
ArtEq & 2.615 & 3.750 & 3.054 & 3.426 & 2.017 & 4.900\\
\cmidrule(lr){1-1}
Latent-VaRefine & 2.329 & 2.917 & 2.463 & 3.216 & 1.729 & 4.533\\
Improvement (\%) & 10.9 & 22.2 & 19.4 & 6.1 & 14.3 & 7.5 \\
\cmidrule(lr){1-1}
Vertex-VaRefine & \textbf{1.447} & \textbf{1.653} & \textbf{1.437} & \textbf{1.627}&\textbf{ 1.100} & 3.149\\
Improvement (\%) & 44.7 & 55.9 & 52.9 & 52.5 & 45.5 & 35.7 \\
\cmidrule(lr){1-1}
ElaOpt & 1.983 & 3.217 & 2.564 & 2.911 & 1.622 & 3.327 \\
Improvement (\%) & 24.2 & 14.2 & 16.0 & 15.0 & 19.6 & 32.1 \\
\cmidrule(lr){1-1}
Latent-VaRefine + ElaOpt  & 1.695 & 2.464 & 1.938 & 2.633 & 1.177 & \textbf{2.812} \\
Improvement (\%) & 35.2 & 34.3 & 36.5 & 23.1 & 41.6 & 42.6 \\
\cmidrule(lr){1-1}

Latent-VaRefine + Vertex-VaRefine & 1.665 & 1.875 & 1.736 & 1.784 & 1.438 & 3.459 \\
Improvement (\%) & 36.3 & 50.0 & 43.2 & 47.9 & 28.7 & 29.4 \\

\midrule

Dataset & EYES\_JAPAN & Human4D & KIT & MPI\_HDM & MPI\_Limits & MPI\_mosh \\
\midrule
ArtEq & 5.400 & 3.774 & 2.595 & 3.994 & 4.299 & 2.849\\

\cmidrule(lr){1-1}
Latent-VaRefine & 4.356 & 3.402 & 2.083 & 3.047 & 4.045 & 2.429\\
Improvement (\%) & 19.3 & 9.9 & 19.7 & 23.7 & 5.9 & 14.7\\

\cmidrule(lr){1-1}
Vertex-VaRefine & \textbf{2.854} &\textbf{ 1.912 }& 1.555 & 2.091 & 2.864 & \textbf{1.400}\\
Improvement (\%) & 47.1 & 49.3 & 40.1 & 47.6 & 33.4 & 50.9 \\

\cmidrule(lr){1-1}
ElaOpt & 4.541 & 3.102 & 2.178 & 3.463 & \textbf{2.449} & 2.155 \\
Improvement (\%) & 15.9 & 17.8 & 16.1 & 13.3 & 43.0 & 24.4 \\

\cmidrule(lr){1-1}
Latent-VaRefine + ElaOpt & 3.488 & 2.563 & 1.658 & 2.826 & 2.451 & 2.194 \\
Improvement (\%) & 35.4 & 32.1 & 36.1 & 29.2 & 43.0 & 23.0 \\

\cmidrule(lr){1-1}
Latent-VaRefine + Vertex-VaRefine & 3.127 & 2.304 & \textbf{1.467} & \textbf{2.066} & 3.322 & 1.813 \\
Improvement (\%) & 42.1 & 39.0 & 43.5 & 48.3 & 22.7 & 36.4 \\

\midrule
Dataset & SOMA & Transitions & BML\_handball & WEIZMANN & TotalCapture & SSM \\
\midrule

ArtEq & 2.650 & 2.940 & 2.667 & 2.989 & 3.244 & 4.321\\

\cmidrule(lr){1-1}
Latent-VaRefine & 2.427 & 2.200 & 2.187 & 2.192 & 2.290 & 2.601\\
Improvement (\%) & 8.4 & 25.2 & 18.0 & 26.7 & 29.4 & 39.8\\

\cmidrule(lr){1-1}
Vertex-VaRefine & \textbf{1.271} & \textbf{1.073} &\textbf{ 1.419} & \textbf{1.152} & \textbf{1.123} & \textbf{1.647}\\
Improvement (\%) & 52.0 & 63.5 & 46.8 & 61.5 & 65.4 & 61.9 \\
\cmidrule(lr){1-1}
ElaOpt & 2.257 & 2.209 & 2.029 & 2.587 & 2.950 & 3.877 \\
Improvement (\%) & 14.8 & 24.9 & 23.9 & 13.4 & 9.06 & 10.3 \\

\cmidrule(lr){1-1}
Latent-VaRefine + ElaOpt & 1.961 & 1.431 & 1.757 & 1.741 & 1.634 & 1.837 \\
Improvement (\%) & 26.0 & 51.3 & 34.1 & 41.8 & 49.6 & 57.5 \\

\cmidrule(lr){1-1}
Latent-VaRefine + Vertex-VaRefine & 1.599 & 1.301 & 1.512 & 1.491 & 1.447 & 1.893 \\
Improvement (\%) & 39.7 & 55.7 & 43.3 & 50.1 & 55.4 & 56.2 \\

\bottomrule

\end{tabular}
\end{minipage}%
}%
}
\end{table*}

\begin{table*}[]
\centering
\setlength{\fboxsep}{3pt}
\setlength{\fboxrule}{0.6pt}
\makebox[\textwidth][c]{%
\fbox{%
\begin{minipage}{\textwidth}
\centering
\small
\setlength{\tabcolsep}{2pt}
\renewcommand{\arraystretch}{0.9}

\caption{Detailed evaluation on additional AMASS sub-datasets (SMPL target). We report \textbf{asymmetric chamfer distance} (in cm) for ArtEq, Latent-VaRefine, ElaOpt, and Vertex-VaRefine together with the corresponding relative improvements.}
\label{tab:amass_subdatasets_supp_ACham}
\begin{tabular}{l|cccccc}
\toprule
Dataset & ACCAD & BML\_rub & CMU & CNRS & DFAUST & DanceDB \\
\midrule
ArtEq & 1.409 & 1.955 & 1.645 & 1.443 & 1.245 & 2.627\\

\cmidrule(lr){1-1}
Latent-VaRefine & 1.302 & 1.605 & 1.374 & 1.396 & 1.119 & 2.436\\
Improvement (\%) & 7.59 & 17.9 & 16.5 & 3.26 & 10.1 & 7.27\\

\cmidrule(lr){1-1}
Vertex-VaRefine & 0.893 & 0.987 & 0.875 & 0.898 & 0.776 & 1.730\\
Improvement (\%) & 36.6 & 49.5 & 46.8 & 37.8 & 37.7 & 34.1 \\

\cmidrule(lr){1-1}
ElaOpt & 0.705 & 0.876 & 0.763 & \textbf{0.703} &  0.637 & 0.898\\
Improvement (\%) & 50.0 & 55.2 & 53.6 & 51.3 & 48.8 & 65.8\\

\cmidrule(lr){1-1}
Latent-VaRefine + ElaOpt &  \textbf{0.674} &  \textbf{0.783} & \textbf{0.710} & 0.707 &  \textbf{0.604} & \textbf{0.804}\\
Improvement (\%) & 52.2 & 59.9 & 56.8 & 51.0 & 51.5 & 69.4\\

\cmidrule(lr){1-1}

Latent-VaRefine + Vertex-VaRefine & 0.983 & 1.082 & 1.018 & 0.945 & 0.926 & 1.918 \\
Improvement (\%) & 30.2 & 44.7 & 38.1 & 34.5 & 25.6 & 27.0 \\

\midrule
Dataset & EYES\_JAPAN & Human4D & KIT & MPI\_HDM & MPI\_Limits & MPI\_mosh \\
\midrule

ArtEq & 2.638 & 1.927 & 1.375 & 1.853 & 2.089 & 1.606\\

\cmidrule(lr){1-1}
Latent-VaRefine & 2.182 & 1.753 & 1.194 & 1.504 & 1.933 & 1.395\\
Improvement (\%) & 17.3 & 9.0 & 13.2 & 18.8 & 7.5 & 13.1\\

\cmidrule(lr){1-1}
Vertex-VaRefine & 1.410 & 1.084 & 0.947  & 1.091 & 1.396 & 0.917\\
Improvement (\%) & 46.6 & 43.7 & 31.1 & 41.1 & 33.2 & 42.9 \\

\cmidrule(lr){1-1}
ElaOpt & 1.112 & 0.901 & 0.713 & 0.818 & \textbf{0.711} & 0.682\\
Improvement (\%) & 57.8 & 53.2 & 48.1 & 55.9 & 66.0 & 57.5\\

\cmidrule(lr){1-1}
Latent-VaRefine + ElaOpt & \textbf{0.680} & \textbf{0.776} & \textbf{0.658} & \textbf{0.735} & 0.717 &  \textbf{0.680}\\
Improvement (\%) & 74.2 & 59.7 & 52.1 & 60.3 & 65.7 & 57.7\\

\cmidrule(lr){1-1}
Latent-VaRefine + Vertex-VaRefine & 1.542 & 1.225 & 0.900 & 1.101 & 1.600 & 1.078 \\
Improvement (\%) & 41.5 & 36.4 & 34.5 & 40.6 & 23.4 & 32.9 \\

\midrule
Dataset & SOMA & Transitions & BML\_handball & WEIZMANN & TotalCapture & SSM \\
\midrule
ArtEq & 1.532 & 1.664 & 1.524 & 1.664 & 1.631 & 2.308\\

\cmidrule(lr){1-1}
Latent-VaRefine & 1.418 & 1.294 & 1.255 & 1.379 & 1.289 & 1.605\\
Improvement (\%) & 7.4 & 22.2 & 17.7 & 17.1 & 21.0 & 30.5\\

\cmidrule(lr){1-1}
Vertex-VaRefine & 0.849 & 0.742 & 0.891 & 0.805 & 0.765 & 1.045\\
Improvement (\%) & 44.6 & 55.4 & 41.5 & 51.6 & 53.1 & 54.7 \\

\cmidrule(lr){1-1}
ElaOpt & 0.728 & 0.752 & 0.668 & 0.823 &  0.797 & 1.034\\
Improvement (\%) & 52.5 & 54.8 & 56.2 & 50.5 & 51.1 & 55.2\\

\cmidrule(lr){1-1}
Latent-VaRefine + ElaOpt  & \textbf{0.700} & \textbf{0.651} &  \textbf{0.651} & \textbf{0.714} & \textbf{0.671} & \textbf{0.685}\\
Improvement (\%) & 54.3 & 60.9 & 57.3 & 57.1 & 58.9 & 70.3\\

\cmidrule(lr){1-1}
Latent-VaRefine + Vertex-VaRefine & 0.984 & 0.855 & 0.923 & 0.931 & 1.317 & 1.184 \\
Improvement (\%) & 35.8 & 48.6 & 39.4 & 44.1 & 19.3 & 48.7 \\

\bottomrule
\end{tabular}

\end{minipage}%
}%
}
\end{table*}


%spare a file merge



\end{document}